\documentclass[11pt]{article}
\usepackage{eamt26}
\usepackage{times}
\usepackage{url}
\usepackage{latexsym}
\usepackage[small,bf]{caption} % added MLF 20171211
\usepackage[colorinlistoftodos]{todonotes}
\usepackage{multirow}
\usepackage{multicol}
\usepackage{pdflscape}
\usepackage{tabularx}
\usepackage{makecell} 
\usepackage{amsmath}
\usepackage{booktabs}
\usepackage{tablefootnote}
\usepackage[table]{xcolor}
\usepackage{pdflscape}
\usepackage{enumitem}

\title{Creativity Bias: How Machine Evaluation\\ Struggles with Creativity in Literary Translations}

\author{
  \textbf{Kyo Gerrits}, 
   \textbf{Rik van Noord},
  \textbf{Ana Guerberof Arenas}\\
  Centre for Language and Cognition, University of Groningen \\
  \texttt{k.gerrits@rug.nl}
}

\date{}

\begin{document}
\maketitle
\begin{abstract}
%\todo{AGA: I think the title should have Creativity Bias so the concept is created. If the reviewers think is confusing they will say it}
This article investigates the performance of automatic evaluation metrics (AEMs) and LLM-as-a-judge evaluation on literary translation across multiple languages, genres, and translation modalities. The aim is to assess how well these tools align with professionals when evaluating translation %quality, 
creativity (creative shifts \& errors) %and errors,
and see if they can substitute laborious manual annotations. A dataset of literary translations across three modalities (human translation, machine translation, and post-editing), three genres and three language pairs was created and annotated in detail for %errors and 
creativity by experienced professional literary translators. The results show that both AEMs and LLM-as-a-judge evaluations correlate poorly with professional evaluations %, particularly 
on creativity, with LLM-as-a-judge showing a systematic bias in favour of machine-translated texts and penalising creative and culturally appropriate solutions. Moreover, performance is consistently worse for more literary genres such as poetry. This highlights fundamental limitations of current automatic evaluation tools for literary translation and the need to create new tools that do not frequently consider out of routine translations as errors.

\end{abstract}

\section{Introduction}
As machine translation (MT) has continued to improve significantly in recent years, literary MT is used more and more and its implementation in publishing houses is now a reality \cite{Klemin,Bookseller}. %has become the focal point of numerous studies \rik{This sentence needs a rewrite, it doesn't really flow}
This also creates an increased focus on evaluating literary MT. Within the MT community, automatic evaluation metrics (AEMs) are standard tools for assessing translation quality \cite{schmidtova-etal-2024-automatic-metrics}. Recently, LLMs have also been used as judges for detailed translation annotation \cite{fu-etal-2024-gptscore,wang-etal-2023-chatgpt,zhang-etal-2025-crowd}. 
Professional annotation and quality assessment is still seen as the gold standard, but it can be expensive and time-intensive \cite{chaganty-etal-2018-price}. It would save costs and labour if such annotation could be performed automatically.

Although these metrics are widely adopted within the NLP community as proxies for  translation quality \cite{lavie-etal-2025-findings,schmidtova-etal-2024-automatic-metrics}, translation scholars have expressed critique and doubt about their usefulness and validity:
they argue that these metrics are designed to capture superficial similarity to the source or a reference, and penalise all shifts and deviations, even when these shifts are exactly what makes a translation creative, culturally suited and literary accomplished \cite{zhang-etal-2025-good,blagec-etal-2022-global,Way2018}.
Our previous study analysing creativity in MT output from LLMs \cite{du-etal-2025-optimising} found little correlation between AEMs and human annotation, but this was not the main focus of the study. Here we would like to address this issue in more depth. %\rik{Hmm this is unusual in an NLP publication. You would usually say you build on the work of Kyo et al (2025) without saying it is you} 
%we only focused on one relatively small text. We want to explore this further. \todo{KG: I'm not very happy with the final sentence and not sure if this makes sense now.}

In this paper, %This paper explores these two viewpoints in more detail.
we report on an analysis of automatic metric performances across multiple literary translations, including three language pairs, genres and translation modalities (human translation (HT), post-editing (PE) \& MT). We compare these to fine-grained annotations for creativity (errors \&  creative shifts) %creativity
from experienced professional literary translators. Specifically, we ask three questions: %with the same subquestion for both:

\vspace{0.1cm}
\noindent \textbf{RQ1:} How well do automatic evaluation metrics align with professional evaluation in literary translation?

\vspace{0.1cm}
\noindent \textbf{RQ2:} How well does LLM-as-a-judge evaluate creativity in literary translation (creative shifts \& errors), compared to professionals? 

\vspace{0.1cm}
\noindent \textbf{RQ3}: Do these results change across genres? %NOG CONTRIBUTIONS TOEVOEGEN

%Our paper focuses on how automatic metrics and annotations relate to human annotations on literary translation across modalities, genres and language pairs. First we will consider previous work related to our current study, in terms of literary MT, human annotations of error and creativity and automatic evaluation systems. Then we will discuss our set-up, including the texts we use (ST, TT, genre), how the translations were done (both human and MT), how the annotations were created (again both human and MT) and how to compare these annotations. We can then move on to our results and our discussion about those results. %Not sure whether I like this section like this here, but I think for now it helps structure the paper a little 

\vspace{0.1cm}
\noindent Our main findings and contributions are: %\todo{KG: Do we need both findings and contributions? AGA: I think so, yes.}
\begin{enumerate}
    \item We introduce a \textbf{dataset} of literary translations across 2 source languages, %(English \& Russian), 
    2 target languages%(Dutch \& Catalan)
    , 3 genres%(poem, short story \& thriller)
    , and 3 modalities (HT, MT, PE). 
    HT and PE are created by professional literary translators %(with 10+ years experience in literary translation) \rik{The stuff in parentheses can be removed} 
    and all texts are annotated in detail for creativity (errors and creative shifts). %(following a taxonomy explained in Section \ref{sec:hum_ann})\rik{Same here, shorter is better, no need for the info in parentheses} 
    %by the same four translators and 3 other language experts.% \todo{AGA: Other? How many?KG: I know being precise is important, so we could say 3 other language experts (one per pair), but I added this more as to not be untruthful if we say only the translators, but the point is more that the annotations are annotated by knowledgeable people rather than the amount of language experts. But if we want we can still add the number. AGA: I think you need to add the number}
    \item \textbf{Automatic evaluation metrics correlate negatively to weakly with professional annotations} on creativity in literary translation. %with some metrics even correlating negatively on creative translation solutions
    %suggesting the more creative a translation is, the lower the score. 
    %\item Pre-trained metrics perform equally to string-based ones, with \textbf{COMET even performing worst} of all reference-based metrics. Although reference-based metrics do outperform reference-free ones, all metrics correlate poorly with human judgments. [I took this one out as I think it might be less relevant] 
    \item \textbf{LLM-as-a-judge struggles to identify errors}, both ignoring major errors and including non-errors. It also marks significantly fewer errors in MT and significantly more in HT compared to professional annotation. 
    \item \textbf{Both AEMs and LLM-as-a-judge ignore or penalise creativity in translation}. LLM-as-a-judge marks creative solutions as errors, while AEMs correlate weakly to negatively with creative shifts and both underperform on highly literary and creative genres such as poems compared to thrillers. 
    %We suggest that \textbf{using AEMs or LLM-as-a-judge to evaluate the quality and creativity in the translation of literary works is at present insufficient when compared to professional evaluations.} %More research using more texts across genres and more annotators is needed to study how these metrics deal with creativity in literary translation in more detail.
\end{enumerate}

\section{Related work}
We will first cover recent findings on MT-quality for literary texts in WMT, then we will look at how creativity in translation has been operationalised and quantified, and lastly we will move to research on AEMs and LLM-as-a-judge evaluation. 
 
\subsection{Literary translation and MT}
Recent WMT findings indicate that MT quality for literary translation of top-performing systems is comparable to human translated texts for multiple languages \cite{kocmi-etal-2025-findings}. %with best-performing systems scoring between 87 and 98.6 (on a scale of 100).\rik{Maybe Ana likes this, but for me the scores are not needed here} [I've now put it in percentages, if Ana wants to keep it we can keep it in otherwise I'll remove it.]
Yet the literary texts used were extracted from a fanfiction website; fanfiction adheres to different rules than literature (in terms of characterisation, plot and narrative construction) \cite{FanFic_1,Fanfic_2}, and texts might be written by non-native authors or GenAI \cite{Alfassi04032026}. Furthermore, studies analysing literary MT in more detail continue to find issues, especially those studies focusing on certain literary features: transferring meaning and especially idiomatic phrases and culturally charged expressions continue to be difficult \cite{matusov-2019-challenges,karpinska-iyyer-2023-large,CorpasPastor2024}.
 %We explore the tension between these two points by comparing the quality of different modalities across genres and language pairs. 

%This tension is similar to the tension between quality assessment described above, and we want to explore the tension between these two points by comparing the quality of different modalities across genres and language pairs

%WMT has not yet included any of our language pairs for literary translation (EN-NL, EN-CA, \& RU-NL, see Section \ref{sec:Texts}) \AGA{I think some of these languages were covered in other WMTs, this is the way WMT works, they cannot cover all the languages every year}, but studies analysing literary MT for EN-NL texts show that between 56\% and 77\% of MT-sentences contain errors (Tezcan et al., 2019; Webster et al., 2020, respectively). The latter study hypothesises MT often struggles with linguistic complexity in source texts, which is especially important for literary texts. Guerberof-Arenas and Toral (2020; 2022) also show that MT contains significantly more errors than HT for EN-NL and EN-CA texts. \AGA{I think these articles might be becoming too old and we need to cite newer papers that use more up to date systems. Most of these are NMT. Check papers that work with LLMs, even if older models Like Zhang 2025 or or Macken 2024, Karpynska en Iyyer 2023 and more}

Post-editing (PE) can be seen as a solution for MT issues: it helps remove MT-errors, while being less time-consuming and expensive compared to HT in some cases \cite{castaldo-etal-2025-extending,Tewari2026}, but this is not always found \cite{Terribile2024}. Nevertheless, PE %is often of a lower quality 
presents multiple issues when compared with HT: PE reduces authorial literary voice \cite{KennyWinters,Mohar2020}, and MT can prime post-editors, retaining errors and stylistic features of MT, making PE more like MT than HT \cite{macken-etal-2022-literary,daems-etal-2024-impact}.
%We expect that for our texts too MT has significantly more and more severe errors and that PE has fewer errors and more creative solutions than MT, though still less than HT. We want to see if the AEMs and LLM-as-a-judge repeat this pattern or if they deviate for errors, creativity or both. 

\subsection{Creativity and creativity annotations}
 \label{sec:crea}
Creativity in translations concerns something both new and appropriate: it is a skill often highlighted when referring to literary translation, as translators need to go beyond the word equivalence %or even meaning
to create a text that reflects the voice and intent of the author in another culture \cite{Bassnettetal,Rojo2017,Kaufman01012012}. %\rik{Again, this sounds like a claim where we want to at least cite a paper, or a book on literature translation, something} %a core feature of literary translation and thus needs to be taken into account when evaluating. %We will therefore analyse how well AEMs correlate with creativity and how well LLM-based models annotate creativity in literary translation. 
%However, it is difficult to operationalize creativity and create a clear framework to identify it. %clearly and even more so to have a clear framework for its annotation. 
Some researchers have operationalised creativity by using a framework for manual annotation. For example, Guerberof-Arenas and Toral \shortcite{Guerberof2020}  measure creativity in literary translations by comparing professional translations with MT outputs in literary texts (see Section \ref{sec:hum_ann}). This framework is based on earlier research by Kussmaul \shortcite{Kussmaul1991,Kussmaul1995,Kussmaul2000a} and Bayer-Hohenwarter, who  operationalises creativity %, as part of the TransComp group, 
to analyse the differences between students and professionals' translations \shortcite{Bayer-Hohenwarter2009,Bayer-Hohenwarter2011,Bayer-Hohenwarter2013}.

MT outputs consistently show lower creativity than professional translations \cite{Guerberof2020,Guerberof2022,CorpasPastor2024}. %We also expect this to be the case in this study and we are also interested to see how the automatic metrics will evaluate creativity in MT texts.  %for our texts\AGA{but we have not talked about our texts yet, so perhaps in our experimental setting}.
Research analysing creativity in automatically generated texts (so non-translations) using human judges or creativity frameworks also finds lower creativity scores for machine-generated texts when compared to texts created by writers \cite{belouadi-eger-2023-bygpt5,Chakrabarty2024,chen-etal-2024-evaluating-diversity}. We expect lower creativity for our MT-texts, and we are interested to see whether AEMs and LLM-as-a-judge also reflect this.

%\rik{I feel like this paragraph needs a bit more interpretation: why are we telling the reader this?}  

%annotated creativity index will compare to automatic evaluations %of creativity and whether LLM-based models can calculate CIs automatically in different MT outputs %also find these differences between modalities. \AGA{Is this what you meant?}.

\subsection{Automatic evaluation: AEM and LLM-as-a-judge}
%\AGA{All of these are AEMs but with different methods}.
\label{sec:AEMs}

%AEMs output a single score while LLM-based evaluation is more detailed and outputs features as errors or creative segments depending on the prompt.
%\AGA{But I am not sure that this distinction is real. AEM means automatic evaluation metrics, so the AEM can have an LLM behind it, right?}.

AEMs are often based on the notion that the more similar a translation is to the reference--through n-gram overlap or learned representations that capture semantic, syntactic or contextual proximity--the higher the score. AEMs can be very useful to test MT systems during training, but their validity is less %obvious
pertinent when applied to literary translations where different creative solutions are possible and stylistic divergences, cultural shifts and literary features resist a single interpretation and therefore a single translation \cite{zhang-etal-2025-good}. 
%In other words, there is not one correct translation, and multiple highly different though valid ones can exist simultaneously which problematises the underlying assumption of such metrics. 
BLEU \cite{Papineni2002}, for instance, has been criticised for its over-reliance on strict ordering of surface elements as early as 2006 \cite{callison-burch-etal-2006-evaluating}. Other string-based methods like TER \cite{snover-etal-2006-study} and chrF \cite{popovic-2016-chrf} also do not correlate well with professional scores \cite{novikova-etal-2017-need,mathur-etal-2020-tangled}. 

Pre-trained models like COMET \cite{rei-etal-2020-comet}, BERTscore \cite{zhang2020bertscoreevaluatingtextgeneration} and BLEURT \cite{sellam2020bleurt} outperform string-based metrics \cite{freitag-etal-2022-results,thai-etal-2022-exploring}. Of these, COMET is used frequently and often outperforms others \cite{amrhein-etal-2022-aces,zouhar-etal-2024-fine,wu-etal-2024-evaluating-automatic}. Still, pre-trained metrics also fail to differentiate between critical and non-critical errors \cite{saadany-orasan-2021-bleu} and stylistic or nuanced differences \cite{Mukherjee2025,agrawal-etal-2024-automatic-metrics,hanna-bojar-2021-fine}. They also struggle more at segment-level than at document-level  %whichwe also use
\cite{moghe-etal-2023-extrinsic,freitag-etal-2023-results}.\footnote{However, we use document-level as segment-level created little output, for Limitations see Appendix \ref{sec:limitations}.} %Kocmi et al. \shortcite{kocmi-etal-2021-ship} suggest pairwise rankings work best to decide the better system. %, as we also explore with correlations.
Zhang et al. \shortcite{zhang-etal-2025-good} %\rik{Why the ref like this?},
showed that even recent SOTA metrics prefer machine-generated translations above those translated by professionals. 
%We are also interested to see whether we replicate this for literary translation and whether the LLM-based models also show similar preferences.

%Besides AEMs,
Recently, LLMs are also used for more fine-grained evaluation of translations through prompting. LLM-as-a-judge returns error spans, categories and severity, which helps identifying problematic segments and facilitates improving or post-editing texts \cite{zouhar-etal-2025-ai,xu-etal-2023-instructscore} while also giving a better understanding of how LLMs analyse errors. Most prompting strategies have similar bases and focus on error evaluation using the MQM-framework \cite{lommel2024multirangetheorytranslationquality}, including error categories and severity weights. However, to our knowledge, these evaluations have not been tested for literary translation, as presented in this paper. % We want to see how well they can evaluate such literary texts and how they deal with errors and creativity in literary translation. To do so, we employ the prompt from Tagged Span Annotation (TSA) \cite{yeom-etal-2025-tagged}, which outperformed other techniques in our evaluation.
%\footnote{For more on this evaluation, see Appendix \ref{sec:GPT-evaluation}.}  %We will also consider whether these models prefer machine-generated translations above those done by professionals too. [CHECK:IKWEETNIETMEERZEKERWAARDITSTONDLOL]
%\footnote{They also often measure performance on character-level overlap and some only use major errors to evaluate performance \cite{fernandes-etal-2023-devil}, as minor errors are considered purely stylistic \cite{lu-etal-2024-error}. This skews results and is especially important for literary translation, as minor error might not hinder understanding but hinder enjoyment.}.

 %They tend to be reference-free, although some do include this possibility (such as EAPrompt \cite{lu-etal-2024-error} and G-EVAL \cite{liu-etal-2023-g}). Some include a persona ("You are a careful and balanced annotator for machine translation quality") in their prompt %but they tend to be short \AGA{short? what do you mean that the prompt is not extensive?} 

More recently, using LLMs in a multi-agent framework for evaluation has gained more attention. In this paradigm, agents are assigned specific features of evaluation which are then discussed and weighted \cite{feng-etal-2025-mad,kim2025maslitevalmultiagentliterary}. %For instance, HiMate \cite{feng-etal-2025-mad} uses two-tier agents for main category and sub category errors in the MQM framework who decide together on the error and its category. 
We did not incorporate this in our study as multi-agent frameworks introduce considerable additional complexity (system design, prompt engineering, and computational cost) and are therefore beyond the scope of our exploratory research.

%LLMs are recently also used for more fine-grained evaluation of translations through prompting. This can either be done with a LLM-as-a-judge paradigm or multi-agent frameworks. Although the latter shows potential and balances different abilities of LLMs well \cite{feng-etal-2025-mad,kim2025maslitevalmultiagentliterary}, we decided to use the more straight-forward LLM-as-a-judge paradigm, as multi-agent frameworks introduce considerable additional complexity (system design, prompt engineering, and computational cost) and, crucially, lack standardised implementations, making controlled comparison across studies difficult \cite{you2026agentasajudge}. Instead, we analyse single-model LLM-as-a-judge together with established AEMs as baseline. \todo{KG: I've now added this but I'm not sure if it makes sense in this way.}

To our knowledge, there have been no attempts to have LLM-as-a-judge evaluate creativity. Some studies have however attempted to analyse creativity automatically in texts \cite{qiu-hu-2025-deep,Bielova_2025,atmakuru2024cs4measuringcreativitylarge,CorpasPastor2023,Kovalkov2021}, and although these models and methods show potential in recognising creativity, %models
they still struggle to identify creative segments and underperform significantly compared to professionals. The novelty of our approach is that we use a taxonomy of creativity when prompting the LLM-as-a-judge
%In our study, the model only has to decide whether a given UCP is a creative shift (CS) or not (Reproduction or Omission) and not identify creativity from the text itself
 (as explained in Section \ref{sec:GPT-evaluation}), to seek annotation results that are closer to professional annotation on creativity. %We therefore do expect it to perform relatively well, although not on par with human annotation. %We are also interested to see whether LLM-based evaluations assign more or less CSs than humans and if it favour MT above HT and PE when evaluating creativity as well, as it seems to do in AEMs \cite{zhang-etal-2025-litransproqa}. [KG: Not sure whether it makes sense to keep this as we do not focus specifcially on Modality in the Results section there.]

\section{Methodology}
In this section, we will discuss how the texts were selected, the translations and annotations processed and how AEM and LLM-as-a-judge were used in the experiment.
%the text selection, the translations, and set-up of the professional human annotations, AEMs and LLM-as-a-judge evaluation, before discussing how we analyse the results.

\subsection{Texts: language pairs and genres}
\label{sec:Texts}
This study employs six source texts from two source languages (English (EN) and Russian (RU)) into two target languages (Dutch (NL) and Catalan (CA)) from three genres (poems, literary short stories and thrillers) across three modalities (HT, PE, \& MT). The texts are approximately 150 words each, which is short and could limit the results, see Appendix \ref{sec:limitations}. %\footnote{The texts have been taken from an ongoing experimental study into reader reception.} \todo{AGA: Is this footnote relevant for the paper?} 
None of them have been translated and published into these %our
target languages to our knowledge before. An overview of the texts can be found in Table \ref{tab:Texts}. 

\begin{table}[t]
    \centering
    \small
    \renewcommand\cellalign{tl}
    \renewcommand{\arraystretch}{1.3}
    \resizebox{\columnwidth}{!}{
    \setlength{\tabcolsep}{4pt}
    \begin{tabular}{l l l l l}  
    \toprule
    \bf Title & \bf \# Words & \bf Author & \bf SL & \bf Genre \\
    \midrule
    A Bath & 172 & Amy Lowell & EN & \makecell[l]{Poem \\ }  \\
    \makecell[l]{Afternoon in \\ Summer*} & \makecell[l]{182} & {\makecell[l]{Sylvia Town- \\ send Warner}} & EN & \makecell[l]{Short \\ Story}\\
    \makecell[l]{Bright Young \\ Women*} & \makecell[l]{154} &  \makecell[l]{Jessica Knoll} & EN & Thriller\\
     Your house* & \makecell[l]{97} & \makecell[l]{Bella \\ Akhamadulina} & RU & Poem \\
    Yasha's Dream* & \makecell[l]{169} & \makecell[l]{Anna \\ Starobinets}  & RU & \makecell[l]{Short \\ Story} \\
    Night Watch*  & \makecell[l]{151} & \makecell[l]{Sergei \\ Lukyanenko} & RU & Thriller \\
    \bottomrule
    \end{tabular}
    }
    \captionsetup{justification=centering, font=small}
    \caption{An overview of the texts used, including word counts. Russian titles are in English for readability.\\ * indicates that only a fragment was used.}
    \label{tab:Texts}
\end{table}

\paragraph{Genres} The three genres (poem, short story \& thriller) are included to analyse whether automatic metrics perform better on certain genres, specifically whether they perform better on ‘lower' and relatively less creative genres. Research suggests that the impact of MT on creativity and readers is higher for literary texts that favour style over action \cite{Guerberof2020}. 
We also hypothesise that genre-fiction, such as thrillers, is easier to translate for LLMs due to its conventionality \cite{AbdulGhaffar2024}; publishing houses also make this distinction when deciding to use MT to translate their books \cite{creamer_dutch_2024}. High(er) literature, on the other hand, might pose more of an issue for these models as the focus is not only on the narrative but also on style. Poetry, with the tension between form and content, especially represents a %, forms
challenge \cite{SharmaPoetry,chakrabarty-etal-2021}. %Striking the right balance is difficult for LLMs, as they struggle with rhyme and meter in languages other than English \cite{resende-hadley-2024-translators,ZumerRubab}. 
On the other hand, the poetic license afforded to poetry could actually give some allowance to LLMs in coming up with %hallucinations,
unexpected text and language use, or surprising collocations. We analyse whether genre impacts human evaluation %for translation
and specifically whether the automatic metrics also perform differently across these genres.

%analyse differences among literariness and styles. Research by Guerberof-Arenas and Toral (2020) suggested that the impact of MT on creativity and readers is higher for literary texts that favour style over action. 
%than action-driven narratives.

%, as they are more likely to come up with the most conventional phrasing

\paragraph{Languages} We include multiple language pairs to investigate if there are indications of different results for more distant language pairs. We include two distinct source languages (EN \& RU) and two target languages (NL \& CA), one of which is considered a low-resource language (CA).

\subsection{Translations: HT, PE and MT}
\label{sec:Transl}
\paragraph{HT \& PE}The texts were translated by experienced professional literary translators. We contracted two EN-NL and RU-NL translators, who have each %had more than 12 years of experience in both languages and 
 translated more than 15 novels in these language combinations.\footnote{Found via the \textit{Expertisecentrum Literair Vertalen}, a Dutch literary translation database.} %in 'high literature'. 
They each translated half of the texts on their own (HT) and the other half post-edited (PE). This was alternated between languages, so, for example, if one translated the English thriller on their own, then this translator post-edited the Russian thriller (see Appendix \ref{sec:counterbalancingtransl} for details). Balancing languages and modalities across translators attempts to mitigate interpreting individual translator’s effects as effects of language or modality.
For CA, two EN-CA translators were hired to carry out the task. They have translated more than 12 literary novels from English.\footnote{Found via \textit{AELC}, a Catalan literary translation database.} We followed the same process as before, one translator did half of the texts on their own, and post-edited the second half, while the other one followed the opposite order. We did not apply any discounted rate for post-editing.  The RU-CA was carried out by two other translators and is still under evaluation. It is therefore not included in the results.% and we will not include it in the analysis.

%\todo{Reference to this method in the literature. (KG: I'm not sure I understand the first part of this comment. Should I reference that you \& Antonio did this in other studies as well? Or mention in the lit. review that we always make sure to contact experienced literary translators?)} \todo{AGA: I was referring to the article where we do this, so you can cite the article with the page number because I think we have a graphic. KG: I'm not sure which article this is, I cannot find any graphic about this. (I'm happy to cite (one of) the articles, but I'm keeping this comment for now so we can have a look at it together.}

\paragraph{MT} To create our MT output we tried out different systems, including different prompts for the LLMs. This was done to make sure the MT output generated showed high levels of accuracy, fluency and style.
%we used was of the highest quality attainable by the model at the time of creation. 
%\todo{KG: "highest" sounds very dramatic, but I didn't know a clear way to describe this otherwise. AGA: I do not know if this is better}
We tried out multiple different strategies for the prompt, including a persona \cite{he-2024-prompting}, zero-shot or few-shot \cite{castaldo-monti-2024-prompting,zhang-etal-2023-machine,pmlr-v202-zhang23m}, information about the author, author’s style, context or content of the fragment \cite{Opaluwah_2025,Gao2024,cui-etal-2024-efficiently,egdom-etal-2024-make,yamada-2023-optimizing}, or specifying to “translate creatively” \cite{du-etal-2025-optimising}.
Outcomes were evaluated per sentence based on CSs and error, and on translation preference. A zero-shot English-language prompt to translate creatively, including information on author and the context of the fragment in GPT-4o (the then most recent model, March 2025) 
%\rik{Add footnote with the data when ChatGPT4o was used? KG: but there was not one date, I could do a month?} 
worked best across all conditions.\footnote{See Appendix \ref{sec:MT-template} for the prompt template. All prompts and evaluation protocols are available via \url{https://github.com/INCREC/Creativity_bias}.} The MT also formed the basis for the PE task.

%\rik{Indeed, we have to be careful with phrasing here, as discussed in last meeting}

\begin{table*}[h]
   %\small
    \centering
    \resizebox{0.95\textwidth}{!}{
    \begin{tabular}{@{}ccccc@{}}
    \toprule
    \bf ST & \bf UCP & \bf Modality & \bf TT &  \bf Classification \\ \midrule
    \multirow{3}{*}{\shortstack[c]{Sally had emerged\\from her book}} & \multirow{3}{*}{had emerged}& HT & havia aixecat el cap (had lifted the head)  & CS  \\
    && PE & havia emergit (had emerged) & R \\
    && MT & havia sortit (had come out) & R \\
    \bottomrule
    \end{tabular}
    }
    %\captionsetup{justification=centering, font=small}
    \caption{A sample of creativity annotations, with the ST sentence, the UCP in question, the three modalities, the translation in the modalities and their classification. PE \& MT are reproductions (R) as they reproduce the structure, form and meaning of the ST exactly, whereas HT changes the form and makes the movement more explicit, making it a creative shift (CS).}
    \label{tab:sample_crea_ann}
\end{table*}

\subsection{Annotation framework}
\label{sec:hum_ann}

%\begin{table*}[htbp]
   % \scriptsize
    %\centering
    %\begin{tabular}{@{}cc|cc|cc|cc@{}}
    %\toprule
    %\bf ST & \bf UCP & \bf HT & \bf Class.  & \bf PE  & \bf Class.  & \bf MT  & \bf Class. \\
    %\midrule
    %Sally had emerged from her book & had emerged & havia aixecat el cap  & CS & havia emergit & R & havia sortit & R  \\
    %to the place between the coat closet & coat closet & garderobe voor jassen & R & halkast & CS & kapstok & CS\\
    %The day is fresh-wased and fair & fresh-washed & kraakschoon & CS & frisgewassen & R & pas gewassen & R \\
    %\bottomrule
    %\end{tabular}
    %\caption{A sample of creativity annotations.}
    %\label{tab:sample_crea_ann}
%\end{table*}

The professional annotations serve as a gold reference to compare with automatic evaluations. 
%We want to compare automatic evaluations to the evaluation of experienced annotators to see how well automatic systems perform in judging literary translation, specifically for creativity. 
The annotations were based on the framework to measure creativity in literary translations, already mentioned in Section \ref{sec:crea}. This framework focuses on units of creative potential (UCPs): problematic units in an ST that translators cannot translate routinely and for which they have to use problem-solving abilities--their creative skills \cite{Bayer-Hohenwarter2011}. These can be metaphors, colloquial language, idioms, comparisons, etc.\footnote{For more, see Guerberof-Arenas and Toral \shortcite{Guerberof2022}, p.191.} In the TT, these UCPs can be resolved in a number of ways, the most common ones are:

\begin{enumerate}[itemsep=0pt]
    \item Creative Shift (CS)
    \item Reproduction (R)
\end{enumerate}

\begin{figure}[h]
    \centering
    \includegraphics[scale=0.28]{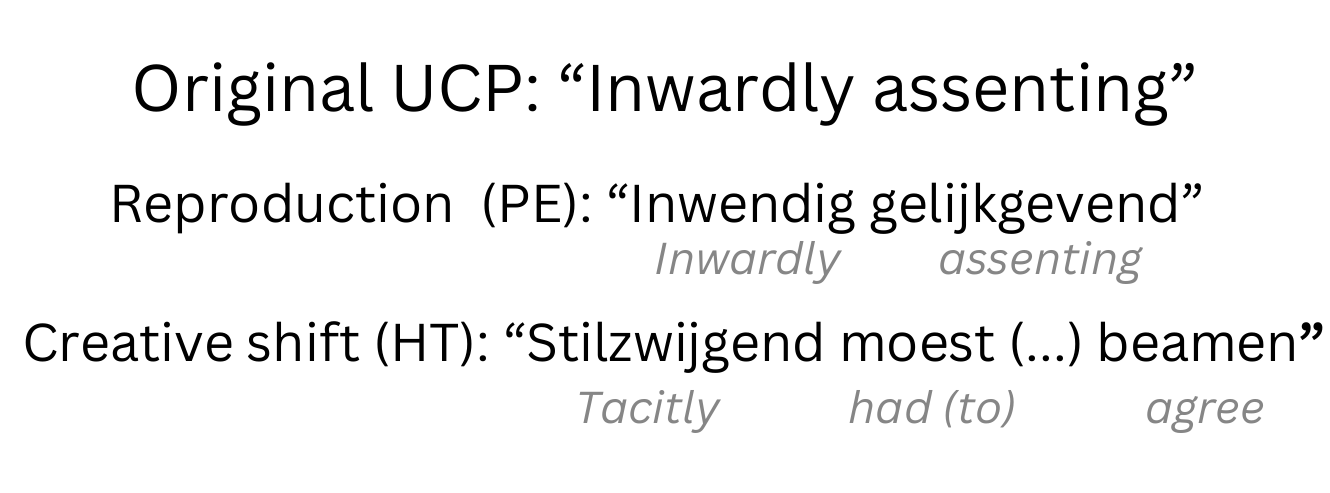}
    \caption{Example of a UCP in the ST with a Reproduction in the PE and a CS in the HT with English glosses underneath.}
    \label{fig:UCP_ex}
\end{figure}

%\rik{See if you like it. It makes CS/R stand out a bit more, and this is an important concept for a reader to see}
CSs are translated UCPs in which the translation deviates from the original. %\footnote{There are multiple types of CSs depending on whether the solution is abstract (CSA), concrete (CSC) or modified in general (CSM). These are all binned as CS. For more on the framework see Guerberof \& Toral (2020).} 
Reproductions, on the other hand, are translations that do not deviate from the original or already-coined translations. A UCP with a reproduction and creative shift is shown as an example in Figure \ref{fig:UCP_ex} for EN-NL: the adverb-gerund subclause in the original is reproduced in PE, whereas HT changes the structure to a main clause, slightly changes the meaning of words and makes it more idiomatic Dutch--in other words, a creative shift.

The creative shift annotations were done by 3 doctoral students who are proficient with the framework that were also translators and were native or proficient speakers for the language pair. They received the texts and the UCP annotations (created by 2 of the doctoral students) and could mark a solution as Creative Shift, Reproduction or Omission. An example of these annotations is shown in Table \ref{tab:sample_crea_ann}.
%\rik{I think the caption of the table should explain what CS and R are. KG: I've added this now but isn't this a bit doubled with Figure 1 then?} %details of the professional annotations can be found in Appendix \ref{sec:detailed_human_ann}. 

Creative solutions should not only be new but also correct. To verify this, professionals also annotated the translations for errors based on the MQM-framework \cite{lommel2024multirangetheorytranslationquality}, including error type %(accuracy, linguistic convention and style) 
and severity. %\footnote{https://themqm.org/error-types-2/typology/}.  
The same professional literary translators who translated the texts also annotated the translations for errors (see Section \ref{sec:Transl}). Annotators could also mark \textit{Kudos} for specifically well-found translation solutions. %These are included in the creativity score below and we want to see how LLM-as-a-judge deals with these. 
Annotations were made through the INCEpTION-platform \cite{klie-etal-2018-inception}.  

%, but switched around. Meaning, if they had done the HT of a given text, they annotated its MT and PE; if they had done the PE of a different text, they annotated its HT.This means the two Dutch translators each annotated 9 texts, while the two Catalan translators annotated 5 texts each.

The annotations for errors and creative shifts are combined to calculate a creativity score, also known as the creativity index \cite{Guerberof2020}:
\vspace{1mm}

$\text{CI} = (\frac{\#CS}{\#UCPs} - \frac{\#ErrorPoints-\#Kudos}{\#Words\_in\_ST}) * 100 $

\vspace{1.8mm}

This score rewards creative solutions, while penalising %(weighted)
error points (errors weighted for severity) based on the number of words and UCPs in the ST. %The score can be used to index creativity in (machine) translations and compare different translations to each other. 
%We want to compare %see how well 
%machine-based CI %scores can index creativity compared
%to the CI by professional annotators.  %and whether their scores are useful to compare translations to each other. 
This model has been used for creativity annotations in literary translation before \cite{castaldo-etal-2025-extending,gerrits-arenas-2025-mt,du-etal-2025-optimising}. 

%by Castaldo et al. \shortcite{castaldo-etal-2025-extending}, Gerrits \& Guerberof-Arenas \shortcite{gerrits-arenas-2025-mt}, and Du et al. \shortcite{du-etal-2025-optimising}, among others. \rik{Why this way of citing?}

%We wanted to know if there were any significant differences across the human error and creativity annotations so we could see if similar patterns arise from the automatic evaluations. 
The professional annotations for the translations used in this paper show that HT was rated significantly higher in creativity %better
than both MT and PE.\footnote{For detailed results, see Appendix \ref{sec:detailed_human_ann}.}  %(fewer errors, more CSs and higher CI).
This is in line with previous research, with HT showing higher creativity than PE, and an even more pronounced difference with MT \cite{CorpasPastor2024,Guerberof2022,Guerberof2020}. 
Looking at genres (for RQ3), we see that the ‘high' literary genres (poems \& short stories) have more CSs and a higher CI than thrillers, which indicates that these ‘higher' genres are also more creative. %We want to see whether the automatic evaluation reflects this.\todo{AGA: I am not sure if this last sentence is necessary} %is of much higher quality than PE and especially MT.

%Genre also had a significant effect on translation creativity %quality
%as poems had significantly more CSs than thrillers (p = .049) and short stories received more Kudos than thrillers, though this failed to reach significance (p = .074). \todo{KG: The Kudos/Genre one is not significant, but I did think it was interesting, but I'm not sure whether it makes sense to include it if it's not significant and if we do how to report on it.} \AGA{It does not make sense. I do not think I would include any of these. I would delete everything after the CI}
%There were no significant differences for errors between the genres (p = .767). High literary genres thus seem to be more creative than lower literary genres as thrillers, though this does not influence error scores. We will see whether similar patterns are present in AEM scores and LLM-based evaluation.\
%\AGA{Delete this section, it is a bit of noise and the paper is very difficult to read as it is. KG: but shouldn't we keep it to say something about genre throughout? RQ3 does focus on genre and I refer back to this finding quite a lot. AGA: If you really like it, insert it in the table and briefly comment on it, and let's see how it reads} 

\subsection{Automatic evaluation metrics (AEMs)}

Our study explores two main types of AEMs, string-based and pre-trained models, besides reference-free and question-based models. An overview of the models used is shown in Table \ref{tab:AEMs}. The first six metrics were run through MATEO \cite{vanroy-etal-2023-mateo}, while the others were run through their GitHub implementations. %\rik{You wouldn't generally say this, more something like through their Github implementations} 
The HT created for this study were used as reference. %\footnote{We use reference-based metrics as they generally outperform reference-free metrics when the reference is of high-quality \cite{freitag-etal-2023-results}.} \rik{Is this footnote needed?}  %But this is not entirely true as LiTransProQA is neither. 

%We include 3 common string-based methods (BLEU \cite{Papieni2002}, TER \cite{snover-etal-2006-study} and Chrf2 \cite{popovic-2016-chrf}) and 3 common pre-trained models (COMET \cite{rei-etal-2020-comet}, BERTscore \cite{zhang2020bertscoreevaluatingtextgeneration} and BLEURT \cite{sellam2020bleurt}). We ran these metrics through MATEO \cite{vanroy-etal-2023-mateo}, using our HT as reference.\footnote{We use reference-based metrics as they generally outperform reference-free metrics when the reference is of high-quality \cite{freitag-etal-2023-results}.}

%Besides these six metrics, we include three new and state-of-the-art metrics, COMETKiwi \cite{rei-etal-2022-cometkiwi}, MetricX24 \cite{juraska-etal-2024-metricx} and LiTransProQA \cite{zhang-etal-2025-litransproqa}. The first two are used in the WMT evaluation tasks \cite{kocmi-etal-2025-findings-wmt25,kocmi-etal-2025-findings,kocmi-etal-2024-findings}, with the first being reference-free QE. LiTransProQA is also reference-free and it is a LLM-based question-answering framework resulting in a (weighted) score based on twenty quality questions about the translation given a source text. The questions are meant to tackle problems literary translators and translations face better than the more surface-based methods employed by string-based metrics and even by pre-trained models. 

\begin{table}[h!]
    \centering
    \small
    \renewcommand\cellalign{tl}
    \renewcommand{\arraystretch}{1.3}
    \resizebox{\columnwidth}{!}{
    \setlength{\tabcolsep}{4pt}
    \begin{tabular}{l l l l }  
    \toprule
    \bf AEM & \bf Type & \bf Ref  & \bf Source \\
    \midrule
    BLEU & string-based & Yes & Papineni et al. \shortcite{Papineni2002}  \\
    TER & string-based & Yes & Snover et al. \shortcite{snover-etal-2006-study} \\
    ChrF2 & string-based & Yes & Popovi{\'c} \shortcite{popovic-2016-chrf} \\
    COMET & pre-trained & Yes & Rei et al. \shortcite{rei-etal-2020-comet} \\
    BERTscore & pre-trained & Yes & Zhang et al. \shortcite{zhang2020bertscoreevaluatingtextgeneration} \\
    BLEURT & pre-trained & Yes & Sellam et al. \shortcite{sellam2020bleurt}  \\
    COMETKiwi & pre-trained & No & Rei et al. \shortcite{rei-etal-2022-cometkiwi} \\
    MetricX24 & pre-trained & Yes & Juraska et al. \shortcite{juraska-etal-2024-metricx} \\
    LiTransProQA & \makecell[l]{question- \\ answering} & No & Zhang et al. \shortcite{zhang-etal-2025-litransproqa} \\
    \bottomrule
    \end{tabular}
    }
    \caption{The AEMs analysed in this paper.}
    \label{tab:AEMs}
\end{table}

\subsection{LLM-as-a-judge evaluation prompting}
\label{sec:GPT-evaluation}
We also wanted to investigate how well LLM-as-a-judge recognises and identifies errors and creative shifts. We let LLM-as-a-judge annotate errors and creative shifts separately, %using the MQM-framework for errors and the UCP-framework for CS. 
which was then combined in an LLM-as-a-judge CI score. 

For the error annotation with LLM-as-a-judge, we opted for Tagged Span Annotation by Yeom et al. \shortcite{yeom-etal-2025-tagged} using GPT-5.2 with high reasoning effort, after trying out different prompting strategies across different models.\footnote{This occurred after creating the MT, which is why we use a newer model. For an overview of our prompting trial, see Appendix \ref{sec:Prompt}, and Appendix \ref{sec:LLM_template} for the prompt.}

For our automatic creative shifts analysis, we employ a similar prompt setup as for the error prompting, with a short persona (“You are a careful and balanced annotator for creativity in translation"), evaluation guidelines and creativity categories. The models were presented with the list of already identified UCPs and classified these, based on the protocol given to the researchers who identified the UCPs (see Section \ref{sec:hum_ann}).\footnote{We tried having the model detect UCPs, but it marked almost all sentences as UCP, even with strict guidelines. Moreover, comparing the model's performance annotating UCPs when different sets of UCP are used would be impossible.} %UCPs were whole sentences covering more than half of texts (averaging 8.7 words/UCP), whereas human-identified UCPs are only 2.7 words.} 
Using the already established list of UCPs we compare the annotations of the model to the annotations of the language experts, both in terms of CS classification and calculated CI score.%\footnote{It would have been impossible to compare performance in annotating UCPs when different sets of UCP are used, which is another reason to use the human-identified list.} \rik{We basically have two footnotes to motivate the same things, can that be merged? And shortened?}
%\AGA{Also, I would include a screen shot with an example. KG: A screenshot of the actual output or our attempt? AGA: Yes, the screenshot of the output by the LLM, similar to Table 2, so we can see Rik: I'd say in the Appendix then, but not sure this is needed}

%\todo{KG: Rik and I discussed perhaps adding a section on this, I did try out some prompts as reported here and indeed there were some issues with this. We did technically however not do this beforehand, as we started with the UCPs already indicated (as Ana had done before). AGA: Do you mean giving the model the list of possible UCPs? KG: No, they were the already decided upon UCPs which I had annotated (together with Nastja).}

%The texts were presented in a .csv-file, in which each line contained one UCP. \AGA{Give us a little graphic with the example. It is so much clearer}Each line also contained the source sentence, the translation of that sentence, the UCP in the ST and the solution found in the TT. The automatic annotations were also created using %(raar werkwoord check)
%ChatGPT's GPT-5.2 model. \AGA{Wasn't this explained before. This section needs a bit of structure so that there is clarity in the prompt given. We can discuss}

\subsection{Comparing professional and computer evaluations}
%\AGA{I would reestructure this section giving more clarity, and avoid explaining what you want to see, this will be in the results. For example, I would start directly with To answer RQ1: write again the question, we did X. For RQ2, we did Y. and to answer RQ3: we did Z. Very simple and clear. Any other thing that you found you will explain in Results}
%To answer our RQs we will compare the scores and output from AEMs and LLM-based evaluation to our human annotations for error and creativity. 

\paragraph{RQ1} To answer how well AEMs align with professional evaluation in literary translation, we correlate the AEM scores with the professional annotations for error points, creative shifts and the combined creativity index (CI) score.
%We want to see if any general patterns emerge about their usability and validity for literary translation evaluation. 

\paragraph{RQ2} To determine how well LLM-as-a-judge for errors and creative shifts align with professional evaluation in literary translation, we first analyse the number of errors from the model and the professionals and see how often they match, including how well they match in categorisation of errors. By looking at the errors %outputted
the model marked we try to understand its workings. We then look at how well LLM-as-a-judge classifies creative shifts in translation compared to professionals, and, finally, we analyse whether professional and automatic CI scores %based on annotations by the model or professionals 
correlate. 

\paragraph{RQ3} To investigate whether these aspects %things
change across genres, we analyse the results for RQ1 and RQ2 across genres in more detail. In professional annotation, thrillers contained less creative solutions and received less Kudos than the other two %'higher'
genres (see Section \ref{sec:hum_ann}), and we would like to know if this pattern is also present in automatic evaluation and whether automatic evaluations perform better on relatively less creative genres.

%It is important to point out that we have a relatively small dataset with a small number of annotators per text (two to four in total, see Section \ref{sec:hum_ann}), which means this study is largely exploratory and any findings need to be further studied using a bigger dataset and more annotators. For more on limitations, see Appendix \ref{sec:limitations}.

\paragraph{Limitations} This %exploratory 
study is not exempt from limitations as mentioned throughout. For more detailed information on these, see Appendix \ref{sec:limitations}.

\section{Results}
The results will be presented in three subsections, each one representing the results for each RQ.

\subsection{%RQ1: 
Alignment of AEMs and Professional Annotations}
\label{sec:results_AEMs}
\begin{table*}[t]
    %\scriptsize
    \centering
    \resizebox{\textwidth}{!}{
    \setlength{\tabcolsep}{5pt}
    \begin{tabular}{@{}cc|c|ccccccccc@{}}
    \toprule
    & & \bf CI & \bf  BERTscore$\uparrow$ &  \bf BLEU$\uparrow$  &  \bf BLEURT$\uparrow$  &  \bf chrF2$\uparrow$  &  \bf COMET$\uparrow$  &  \bf COMETKiwi$\uparrow$ &  \bf LiTransProQA$\uparrow$  &  \bf MetricX24$\downarrow$ &  \bf TER$\downarrow$           \\ \midrule
    \multirow{3}{*}{EN-NL} & HT & \bf 62.2 & & & & & & 76.7 & \bf22.3 & \\
    & PE & 28.9 & \bf 87.2 & \bf 37.8 & \bf 70.5 & \bf 57.7 & \bf 82.4 & \bf 80.0 & 19.2 & \bf 2.66 & \bf 50.4 \\
    & MT & -15.9 & 84.2 & 31.4 & 68.0 & 54.2 & 80.0 & 77.5 & 21.4 & 2.91 & 52.5 \\ \midrule \multirow{3}{*}{EN-CA} & HT & \bf 30.0 & & & & & & 63.6 & 14.6 && \\
    & PE & 15.2 & 79.3 & \bf 16.0 & 48.0 & \bf 36.9 & 67.6 &  65.5 & \bf 22.1 & 5.48 & \bf 81.3 \\
    & MT & -31.6 &\bf 79.9 & 14.6 & \bf 48.3 & 35 & \bf 68.8 & \bf 71.4 & 21.2 & \bf 5.06 & 83.6 \\ \midrule
    \multirow{3}{*}{RU-NL} & HT & \bf 60.1 & & & & & & \bf 65.1 & 20.0 && \\
    & PE & 32.2 & \bf 68.3 & \bf 20.3 & 49.5 & \bf 42.4 & 72.7 & 63.1 & 20.0 & 4.99 & \bf 71.8 \\
    & MT & -50.8 & 68.1 & 14.5 & \bf 50.3 & 40.0 & \bf 73.3 & 63.8 & \bf 20.6 & \bf 4.18 & 76.2 \\ 
 \bottomrule
    \end{tabular}
    }
    \captionsetup{justification=centering, font=small}
    \caption{Overview of all AEM scores per modality for each language pair, with CI for comparison with human evaluations.}
    \label{tab:AEMscores_coll} 
\end{table*}

%\rik{If time allows: order the MT metrics from old to new, instead of alphabetically}
%\paragraph{AEM}
Table \ref{tab:AEMscores_coll} shows the AEM scores for each translation modality per language pair with the highest score per metric bolded for each language pair.\footnote{Details and individual AEM scores are in Appendix \ref{sec:AEMscoredetails}.}
EN-NL PE has consistently higher scores than MT throughout (except for LiTransProQA), which agrees with professional annotations who also rate PE above MT.
%There is agreement across EN-NL scores (as PE has higher scores than MT throughout), which also agrees with professional annotations in terms of ranking. \rik{What do you mean with the last part?} %seem consistently high
%Although scores for EN-NL seem consistently high, less agreement is seen in other texts: EN-CA MT has for instance a lower COMET score than RU-NL PE, but higher scores for BERTscore and COMETKiwi.
Less agreement is seen in the other %texts
language pairs. For example, for EN-CA, BERTscore, BLEURT, COMET, COMETKiwi have lower scores for PE than MT, contrary to the other metrics and professional annotations. For RU-NL, BLEURT, COMET, COMETKiwi, and LiTransProQA also score PE lower than MT.
%: EN-CA MT has for instance a lower COMET score than RU-NL PE, but higher scores for BERTscore and COMETKiwi.
%\footnote{This seems to indicate that AEMs perform worse in low-resource languages, which is not unexpected.}
%Scores also do not agree with human judgment ranking: PE scores better than MT across all language pairs, but this is only reflected in EN-NL. In the others, MT outperforms PE in the majority. 
Furthermore, whereas the difference in CI for PE and MT is pronounced in the professional annotation, this is less evident in the AEMs.
%does not show up in the scores where PE and MT are often close together. 
%Whereas the human CI scores show clear differences between and ranking of the modalities, no such picture emerges from the AEMs. There does not seem to be agreement on scores overall, the differences between AEM scores for the modalities are small, with MT scoring higher than PE on multiple occasions. 
This suggests that AEMs perform worse on low-resource languages and in pairs that do not involve English.

%are in less agreement when it comes to low-resource languages and in language pairs that do not include English than in high resource languages that include English in the pair. \rik{Is it though? Because for EN-CA we also see it, right?}

%As mentioned in Appendix \ref{sec:limitations}, our dataset is small for AEMs to perform optimally, however, these metrics are also known to struggle with stylistic differences \cite{zhang-etal-2025-good}. \rik{I don't like this sentence. Why would AEMs not work on a smaller data set? And what would optimal even be?}
%The scores indeed suggest that AEM scores will not be informative or useful when evaluating these literary translations.

%\todo{KG: potentially, we could add that they seem to score EN-NL higher than Ru-NL , even though CI scores do not indicate such a difference for professional annotation. But I'm not sure whether that is relevant here and/or if we want to add somethign else. I've kept it like this for now and we can discuss on Thursday if any more changes / details are needed/required/interesting.}

%To analyse correlations (or lack thereof) between AEMs and human annotations better, we created a heat map with the Spearman correlations between the AEMs and the features of the human annotations (errors, creativity and the combined creativity index (CI), as shown in Figure \ref{fig:heatmap}. Spearman rank correlations are used as the AEMs have different scales and the unequal spread of data created heteroscedasticity across multiple variables. As Spearman ranks the data for each variable it is also more robust to outliers.

\paragraph{Correlations} We want to see how well AEMs align with professional evaluations on errors, creative shifts (CS) and the combined creativity index (CI).\footnote{We focus on correlations between the creativity and the AEMs, so potential differences across other variables (except for Genre, see Section \ref{sec:Genre}) are beyond our scope.} To do so, we created Spearman correlations between these professional evaluations and the AEM scores across all texts collapsing Modality, Genre and Languages. We used Spearman instead of Pearson as AEM-scores are not linearly spread and have different scales. Spearman is also more robust to outliers. %\todo{KG: One of you deleted this section before, but I've added it back in as you also want me to explain why we use Spearman, so I've put it back in}
%\footnote{Spearman is used as the AEMs have different scales.}\AGA{ This is problematic. You need to explain a bit more how the Spearman was run because you have BLEURT not doing great on Table 4, but here with better correlations. Here you do not run this per language combination, so it could be that some results are confounding others. More info is needed} 
%\todo{KG: I have heat maps for the other IVs as well (next to genre, which is already included in Figure 5. I discussed this with Rik before and we didn't want to have too many figures about this and we thought the genre one was most interesting. I can still add them in an appendix and discuss them but I'm not sure whether it adds something}

\begin{figure}[h]
    \centering
    \includegraphics[scale=0.25]{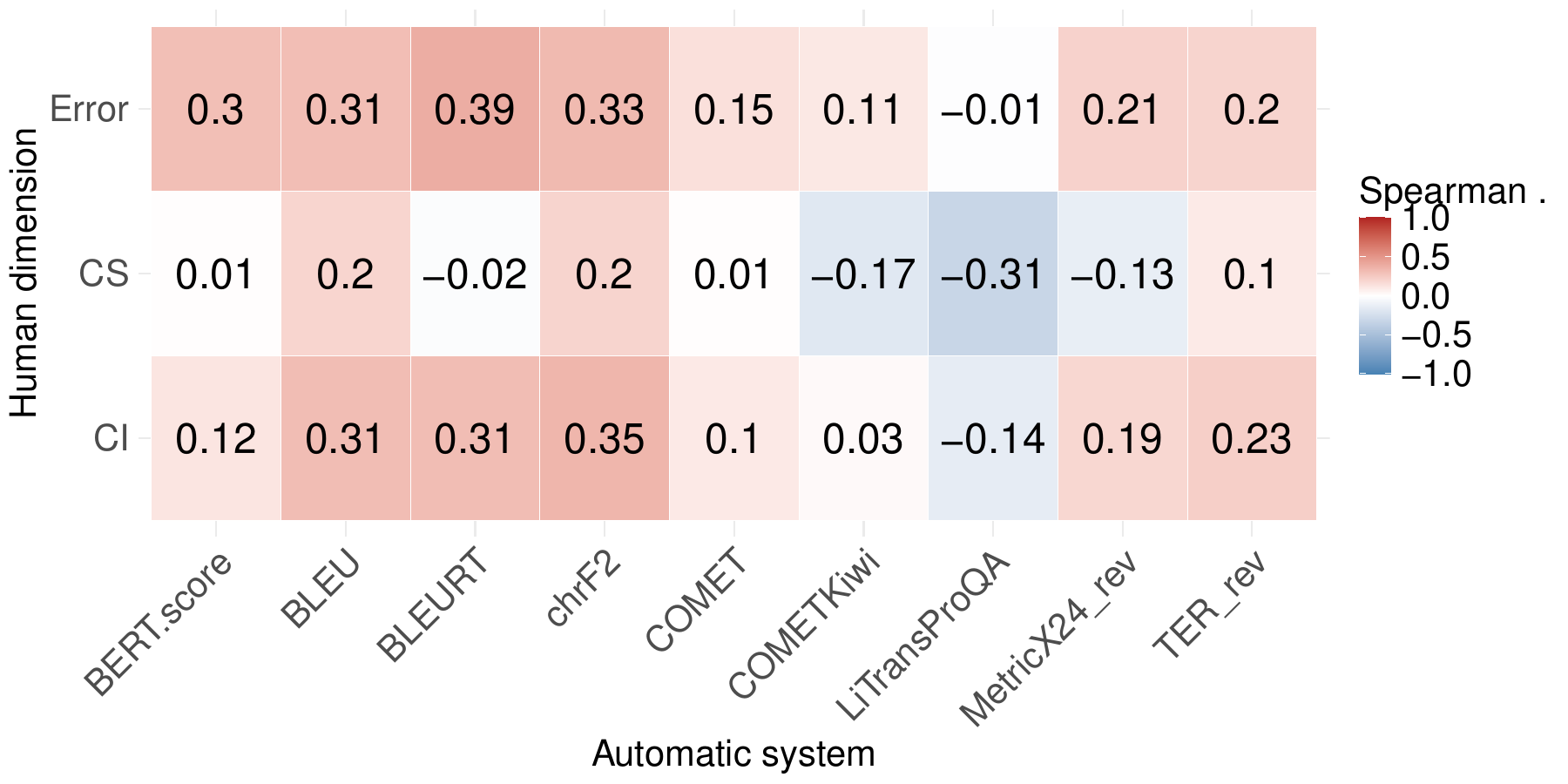}
    \caption{Heat map of the Spearman correlations between AEMs and professional annotations.}
    \label{fig:heatmap}
\end{figure}

The results in Figure \ref{fig:heatmap} show weak correlations. Errors have the strongest correlations (with 0.39 for BLEURT), but even this is moderate to weak.\footnote{Errors are inverted for positive correlations.} For CS, we see weak correlations too, with 4 negative correlations and 0.2 as highest correlation (BLEU \& chrF2). %\footnote{
String-based metrics perform similar to pre-trained models, even though the latter are considered much more reliable \cite{freitag-etal-2022-results}. COMET especially is surprising, as it is considered one of the best AEMs and is frequently used to compare model performance \cite{zouhar-etal-2024-fine,wu-etal-2024-evaluating-automatic}. %}
%For more information, see Appendix \ref{sec:AEM_reference_discussion}\AGA{This Appendix is problematic. %What about Appendix E. This is not even mentioned here, but there is more data.}
The short length of our texts could influence the AEM's scores and our results. Still, AEMs here correlate weakly with errors and disregard or even penalise creative shifts. This lack of correlation is also shown for CI, which is understandable as it rewards creative shifts. 
%do not correlate with creativity and even seem 
%appear to particularly penalise creative shifts. This %problem
%lack of correlation %with creativity
%is also shown %in the correlations
%with CI, and this is understandable because this index rewards creative shifts. Nevertheless, the correlation with errors is also weak. 
%We see that they do not exceed 0.35 (BLEURT), and they are overall low and even negative (LiTransProQA,-0.14).  %This shows that AEMs do not correlate with professional judgments on errors, creative shift or creativity index for these literary translations.

\begin{table}[t]
    %\small
    \centering
        \resizebox{\columnwidth}{!}{
    \setlength{\tabcolsep}{3pt}
    \begin{tabular}{@{}ccc|ccc|ccc|ccc@{}}
    \toprule
        && & \multicolumn{3}{c|}{\bf Poem} & \multicolumn{3}{c|}{\bf Short story} & \multicolumn{3}{c}{\bf Thriller}                             \\ \midrule
         \bf ST & \bf  TT & \bf  From &   \bf HT &  \bf PE & \bf  MT & \bf  HT &  \bf PE &  \bf MT &  \bf HT &  \bf PE &  \bf MT           \\ \midrule
     \multirow{3}{*}{EN}&\multirow{3}{*}{NL} &  Human   & 3 & 8 & 16 & 4 & 5 & 17 & 6 &11 & 9 \\
     & & LLM & 8 & 12 & 12 & 14 & 8 & 12 & 8 & 9 & 9 \\
     && Match & 0 & 3 & 6 &2&0&2&1&4&1 \\
     \midrule 
     \multirow{3}{*}{EN}&\multirow{3}{*}{CA} &  Human & 9 & 7 & 16 & 3 & 7 & 25 & 4 & 7 & 18 \\
     && LLM & 13 & 16 & 9 & 14 & 6 & 8 & 9 & 10 & 5\\
     && Match & 4 & 4 & 7 & 1 & 1 & 4 & 0 & 1 & 4 \\
     \midrule
     \multirow{3}{*}{RU}&\multirow{3}{*}{NL} & Human & 7 & 6 & 14 & 4 & 14 & 14& 9 & 15 & 28 \\
     & & LLM & 19 & 23 & 24 & 10 & 13 & 10 & 9 & 13 &20 \\
     && Match & 4 & 5 & 8 & 2 & 6 & 4 & 0 & 5 & 12\\
 \bottomrule
    \end{tabular}
    }
   %\captionsetup{justification=centering, font=small}
    \caption{Number of errors from professional and machine annotations, including matches.}
    \label{tab:GPTErrors}
\end{table}

\subsection {LLM-as-a-judge}
\label{sec:LLM-as-a-judge}
%\paragraph{Matching errors}

Our second RQ focuses on how well LLM-as-a-judge correlates with professional annotations for creative shifts and errors. %First, we will consider how accurately LLM-as-a-judge marks errors compared to professionals.
Table \ref{tab:GPTErrors} shows the number of errors as annotated by professionals and GPT-5.2, and the number of matches between the two. A match means that the model and professionals identify the same error, but not necessarily that they classify 
%it the same.
the error under the same category.

%nd looking at the actual outputs to find potential causes for its performance. %lelijk verwoord CHECK 
%We will then see how well LLM-based evaluation categorises errors, before moving on to LLM-based evaluation of creativity. Here we will analyse how accurately the model predicts a given UCP before finally correlating the human creativity index (CI) scores with the combined machine scores. [is this too long??].

The low number of matches for all translations shows that LLM-as-a-judge usually does not identify the same errors as the professionals. %, and there seems little difference across language pairs, genres or modalities
 %Although the number of identified errors does not differ largely between professionals and ChatGPT (286 and 323, respectively), there is a clear mismatch in what is marked as error.
Looking at translation modalities, we see that HT is assigned more errors by the LLM-as-a-judge than by professionals, except for the RU-NL thriller (where it receives the same number of errors, but no matches). Similarly, MT is assigned more errors by professionals than by the LLM-as-a-judge, except for RU-NL Poem and all thrillers.\footnote{For more on Genre, see Section \ref{sec:Genre}.} To examine whether these differences across translation modalities are significant, we fitted a binomial generalised linear model (GLM) predicting the source of the error annotation (professional vs LLM). Modality was included as main predictor, with Genre, ST and TT as additional control variables. We find a significant effect of Modality on error markings. Specifically, LLM-as-a-judge is more likely to mark errors in HT (z = 2.07, p = .038), whereas it is less likely to mark errors in MT (z = 6.16, $p<.000$) compared to professionals. This suggests that LLM-as-a-judge overestimates errors in HT and underestimates them in MT, the opposite pattern from professionals.  This also shows that LLM-as-a-judge cannot reliably differentiate between modalities based on error counts, unlike professionals where most errors are marked in MT, followed by PE and then HT. %There are no obvious differences across texts, with both precision and recall performing poorly across the board. This shows that the LLM-as-a-judge is not suited for error assessment in these literary translation and suggests fundamental problems with using LLM-as-a-judge for literary translation evaluation overall.

\begin{table*}[h]
    \centering
    %\scriptsize
    \renewcommand\cellalign{tl}
    \resizebox{\textwidth}{!}{
     \setlength{\tabcolsep}{3pt}
    \begin{tabular}{l l l l}  
    \toprule
    \textbf{ST} & \textbf{Marked as error} & \textbf{Back translation} & \textbf{Explanation} \\
    \midrule
    %\makecell{her artless \\ sophistications} & \makecell{haar argeloze wijsneuzigheden \\ sofismes ingenus} & Both solutions are Kudos marked as error\\
    public library & \makecell{bibliotheek (HT, PE), \\  biblioteca (HT, PE)} & library & \makecell{both commonly used without specifying ‘public', which MT \\ includes  (error for professionals, not for LLM-as-a-judge).}\\
    second floor & eerste verdieping (HT) & first floor & \makecell{US second floor is European first floor (CSI), same phrasing \\ not marked in PE, MT had ‘tweede verdieping' (second floor). } \\
    \makecell{thirteen feet \\ and two inches} & \makecell{quatre metres i deu centímetres (HT, PE)} & 4 meter and 2 centimetres  & \makecell{Metrical shift: EU uses the metric system, not the imperial one, \\ MT retains imperial system and is not marked down by the model.} \\
    fifteen feet & vijf meter (HT, PE, MT)& 5 meter & Metrical shift: EU uses the metric system, not the imperial one.\\
    pours & klatert & splashes & Marked as Kudos by professionals, but as error by model. \\
    inwardly assenting & stilzwijgend moest (...) beamen & tacitly (...) had to agree &  Marked as Kudos by professionals, but as error by model (Figure \ref{fig:UCP_ex}). \\
    %Sally& \textit{la Sally} (HT) && Common language use \\ 
    %\textit{šipučih upsy} & \textit{paracetemol} (HT) & & CSI Common painkillers \\
    \bottomrule
    \end{tabular}
    }
    \caption{Examples of Kudos and culture-specific items (CSI) that are marked as errors by the model.}
    \label{tab:Error_example}
\end{table*} 

%To try and understand why LLM-as-a-judge error performed poorly, 
To explore in depth how the LLM-as-a-judge analyses errors, we look at the specific %we analysed the 
errors it marked. % in detail. 
What stood out was that ‘cultural' creative shifts and segments marked as Kudos by the professional translators were often marked as error (some examples shown in Table \ref{tab:Error_example}). 
%\rik{Also this part is still weak, because we do not really quantify anything, just say that we looked and that something stood out}
Kudos were marked as error 15 out of 38 times they appeared. This was especially visible in EN-NL (9 out of 15 times, 60\%) and EN-CA (5 out of 10 times, 50\%), with 7 even marked as major error. A similar %thing
phenomenon occurs %for culture-specific items
for cultural references: the model penalises %correctly shifted items
creative shifts that are correct, and fails to mark incorrectly translated cultural references. %not incorrectly retained ones. 
This shows that the model fails to distinguish between %erroneous
errors and creative shifts and incorrectly marks creative segments--and thus more creative texts--down when such solutions should rather be promoted. In other words, the model biases against creative solutions.\footnote{More on bias, Stureborg et al. \shortcite{stureborg2024largelanguagemodelsinconsistent} \& Gao et al. \shortcite{Gaoetal}.}
%\rik{Yes I like this explicit interpretation!}

%Also surprising is that these Kudos-as-errors occur more often when the source text was English rather than Russian. Russian STs did not have fewer Kudos than the English ones (1.44 and 1.39 per text on average). The ChatGPT 5-2 model's training was predominantly English, so you would expect better performance in English and not worse (as we will see with categorising errors below). The fact that it then struggles more with Kudos in English than with Russian might further suggest a fundamental issue of the model for evaluation of literary texts. \rik{Too long and also not really a strong point given that there's not so much data, and we don't know anymore what GPT-5.2 was trained on}

%\rik{We really need more explicit structure markers here: subsubsections or paragraph headers. It is too much now}
\paragraph{Categorisation}The last %thing
aspect to explore for the error evaluation was their categorisation according to the MQM framework. Table \ref{tab:ErrorCat} shows a confusion matrix with the distribution of errors per category (Style, Accuracy, Linguistic Convention, and No error) across professional and LLM-as-a-judge annotations. Colours indicate matches: green shows that the model and professionals agreed on both error span and category (there was for instance only 1 error that both mark as Style error), yellow indicates that the model and professionals both marked the error but categorised it differently (for example, both marked \textit{waterbassin} (water basin) as error in the Dutch PE poem, but the model categorised it as an Accuracy error, whereas the professionals marked it as a Style error), red lastly indicates that the error marked by the model or professionals was not marked at all by the other (so professionals mark 88 style errors that the model does not mark at all). In other words: green indicates span and category matches, yellow only span matches and red no match. %\todo{KG: I feel like this has become very long now, so please feel free to shorten but I thought I'd keep it a bit long for now so we can discuss if this makes it clearer. AGA: I think this is finally clear, so well done!). }

\begin{table}[htbp]
    %\scriptsize
    \centering
        \resizebox{\columnwidth}{!}{
    \setlength{\tabcolsep}{4pt}
    \begin{tabular}{@{}cc|cccc| c@{}}
    \toprule
        &&  \multicolumn{4}{c}{\textbf{Professionals}} \\ \midrule
        & & Style & Acc & Ling Con & No error & Total LLM\\ \midrule
     \multirow{4}{*}{\textbf{LLM}}&   Style & \cellcolor[HTML]{05DF72} 1 & \cellcolor[HTML]{FFE72E} 3 & \cellcolor[HTML]{FFE72E} 0 & \cellcolor[HTML]{FF8A9D} 15 & 19  \\
     & Acc & \cellcolor[HTML]{FFE72E} 18 & \cellcolor[HTML]{05DF72}  52 & \cellcolor[HTML]{FFE72E}3 & \cellcolor[HTML]{FF8A9D} 182 & 255 \\
     & Ling Con &\cellcolor[HTML]{FFE72E} 4 & \cellcolor[HTML]{FFE72E} 3 & \cellcolor[HTML]{05DF72} 8 & \cellcolor[HTML]{FF8A9D}36 &51 \\
     & No error & \cellcolor[HTML]{FF8A9D} 88 & \cellcolor[HTML]{FF8A9D} 59 & \cellcolor[HTML]{FF8A9D} 45 & & 192 \\
     \midrule
     \multicolumn{2}{c}{Total human} & 111 & 117 & 56 & 233 & \\
 \bottomrule
    \end{tabular}
    }
    \captionsetup{justification = centering, font=small}
    \caption{Error categorisation. Green shows matches in span \& categories, yellow only span and red no match.}
    \label{tab:ErrorCat}
\end{table}

Of all errors marked by the model, only 92 (the sum of the green and yellow cells) match in terms of span with professionals (28.3\%). Of those span matches, 66.3\% also match the category of the professionals%, which is significantly above chance (z = 8.23, $p < .000$)
. This seems due mainly to accuracy, where 52 error categories match, but the model also overestimates accuracy errors (only 73 (52 + 18 + 3) of the 255 accuracy errors from the model were also marked by professionals). The success rate is much lower for style (only 1 category match out of 4 span matches) and linguistic convention (8 out of 15). So although the model technically performs above chance when categorising errors, this is mainly due to Accuracy which the model assigns incorrectly to many errors. % as well. 
Furthermore, this performance only applies to the small subset of matching error spans which was low to begin with, so its actual performance in error annotation is still poor overall compared to professionals.

%Taken together, this shows that although the model performs above chance when the errors match,
%\footnote{Perhaps providing a model with the list of errors without categories improves performance as also shown for creativity categorisation below. However, it is questionable how useful this will be as errors will have to be marked manually anyway.}\rik{I like that you think about these things, but this seems like an unnecessary footnote} this is mostly due to the overestimation of accuracy errors by the model. 
%This indicates that 
%the model fails to adequately categorise errors according to the MQM-framework when compared to professional annotation. \rik{The last part is very technical, can't we say it in more "normal" language?}
%\rik{I feel like we don't tell the story clear enough here. If I look at Table 7 it seems to me humans and LLMs barely agree on anything. But the text doesn't really drive this message home as much}
%\todo{AGA: The entire Categorisation section is very difficult to read as it does not talk about what we see on the table. KG: I have tried to clarify this further by contrasting span and category matches and make the numbers flow a little more clearly. However, we could also remove this entrie section if that does not help.}
%and is therefore not up to the task of error categorisation in literary translation either.

%Now knowing the LLM-based evaluation model fails to identify and categorise errors correctly, 
\paragraph{CS classification} We are also interested to see if LLM-as-a-judge performs better on categorising creative shifts. As described in Section \ref{sec:GPT-evaluation}, the model classified an already-existing list of UCPs into CS, reproduction or omission. Table \ref{tab:Crea_ann} shows a confusion matrix with the UCP-classification for professionals and LLM-as-a-judge over all texts. The table shows how often professionals and LLMs assigned CS, reproduction (R) and Omission (O) and how well they match. We can see for instance that both marked 90 UCPs as CS, but 110 UCPs were marked by professionals as CS but as reproduction (R) by the model. 

Looking at the table, the model classified 63.6\% (359 out of 564) of the UCPs correctly, shown in the green cells. %To see whether this was significant we ran a binomial generalised linear model,\footnote{Omissions account for only 4\% of the data (25 out of 564), we therefore modelled the annotation as having 50/50 chance of being correct when guessing randomly.} which showed the model's performance to be highly significant (z = 6.57, $p < .000$). \AGA{here again, we have statistics, but it is unclear why or what variables we are looking at}%However, while an effect of 63.6\% is above chance it remains modest to moderate \AGA{According to? Why not leave already 63.5, we understand this clearly}. 
We also see that \protect\mbox{professionals} assign more CSs than LLM-as-a-judge (204 vs 165). We want to know if this difference is significant; as the data is paired (both classified the same set of UCPs) and nominal, we run a McNemar–Bowker test on the entire distribution table. This shows that the classification of UCPs significantly differs between professionals and LLM-as-a-judge ($\chi^2$(1) = 17.15, p = .001), post-hoc pairwise symmetry tests indeed show that the model assigns significantly fewer CSs than the professionals do ($p_{\text{adj}}$ = .012). Looking at translation modalities, LLM-as-a-judge assigns more CSs to MT than {professionals} do (54 vs 35).\footnote{More details on CSs across modalities are in Appendix \ref{sec:app_proportions_crea}.} This could be another indication that LLM-as-a-judge %prefers
favours MT, but we also see that the model assigns very similar numbers of CSs to each modality (HT: 55, PE: 56, MT: 54), which is in contrast with professionals who assign %much
more CSs to HT and fewer to MT. More research into prompting strategies for creative shifts is needed to see if the model indeed favours %prefers
MT or if it simply assigns similar numbers of CSs regardless of the modality. %text.   

%\todo{KG: I'm not very happy with this section. I could remove the part on modalities and keep it in the appendix, as I'm not sure it really tells us anything (if anything it raises more questions that are interesting but not directly relevant to the paper).}
%Furthermore, a proportion test showed that the model assigned CS significantly more to MT compared to humans ($\chi^2$(1) = 11.25, p = .001).\footnote{For more on creative shift classification across modalities, see Appendix \ref{sec:app_proportions_crea}.} So, although the model performs significantly above chance in categorising UCPs, this effect is modest while also being biased for MT and overall assigning reproductions significantly more often than CSs compared to professionals. \rik{This also needs a rewrite, it doesn't flow} This indicates that using LLM-as-a-judge for creative shifts shows some potential, although its modest accuracy and bias towards reproduction in the classification risks underrating %creativity and 
%creative solutions in literary translation, while its bias towards MT (not considering errors) lowers its usability as creativity assessor when compared with professional annotators. \rik{Same here, does it have potential or not?}

\begin{table}[h!]
    %\scriptsize
    \centering
    \renewcommand\cellalign{tl}
     \resizebox{0.8\columnwidth}{!}{
    \begin{tabular}{c c |c c c | c} 
    \toprule
    &&  \multicolumn{3}{c}{\textbf{Professionals}} &\\ \midrule
    & & \# CS & \# R & \# O & Total LLM\\
    \midrule
    \multirow{3}{*}{\textbf{LLM}}& \# CS & \cellcolor[HTML]{05DF72}  90 & 73 & 2 & 165 \\
    & \# R & 110 &\cellcolor[HTML]{05DF72}  265 & 2 & 377 \\
    & \# O & 4 & 14 & \cellcolor[HTML]{05DF72} 4 & 22\\ \midrule
    \multicolumn{2}{c}{Total Professional} & 204 & 352 & 8 & 564 \\ 
    \bottomrule
    \end{tabular}
    }
     %\captionsetup{justification = centering}
    \caption{Distribution of categorisation of CSs, reproductions (R) and omissions (O).}
    \label{tab:Crea_ann}
\end{table}
%\todo{KG: should I also colour yellow and red here? AGA: I think so, especially now since you explained what they mean}

%To see whether the LLM-based evaluation performed better on certain conditions, we ran a glm model for Match using Genre, Modality, ST and TT as predictors. After checking assumptions, the model showed that the odds of matching annotation are 2.3 times higher when the source language is English rather than Russian (z = 3.776, $p < .000$). On the one hand, this is not surprising: most training data of LLMs is in English so it is not remarkable that it performs better on that language. However, we have also seen that the model marked Kudos as errors more often for English than for Russian above. Based on this, we might expect the model to perform worse when annotating creativity for English as well. The fact that this is not the case is puzzling. Perhaps the model can recognise the creativity of the shift well enough when prompted to evaluate creativity only on a set list, while struggling to do so when also looking for errors when these shifts are indiscernible from errors for the model. However, more research into LLM-based evaluations is necessary to clearly make this point. \rik{Needs to be considerably shortened}

\begin{figure}[h]
    \centering
    \includegraphics[scale=0.3]{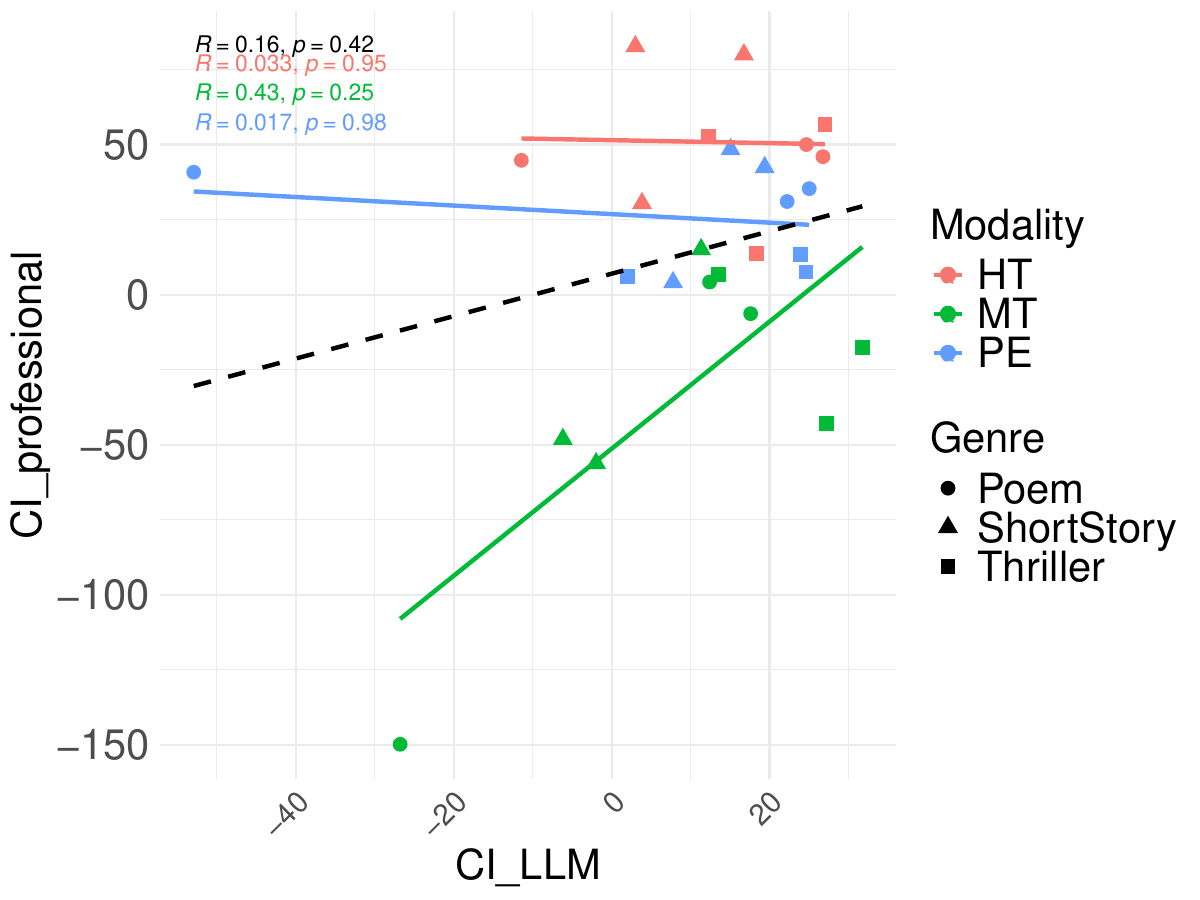}
    \caption{Scatterplot of CI from professional and LLM annotations. Colouring indicates the modality and shapes genre.}
    \label{fig:CIscatter}
\end{figure}

\paragraph{CI scores}Lastly, we want to correlate the CI scores from professionals and LLM-as-a-judge.\footnote{%CI combines annotations for creative shift and error, see Section \ref{sec:hum_ann}. 
Detailed CI per text are in Appendix \ref{sec:CI-scores}.} To analyse this correlation, we created a scatter plot, shown in Figure \ref{fig:CIscatter}.
%we want to see how well combining the error and creativity annotations from LLM-as-a-judge in CI scores correlates to the professional CI scores. \rik{Sorry I think this sentence is too hard to parse for any non-expert reader. What exactly are we correlating and why?}  
The figure shows the CI score from the model (X-axis) and the professionals (Y-axis) for each text. A dashed trend line shows overall correlation, with coloured lines indicating the correlation for each translation modality.\footnote{The shapes reflect genre, see Section \ref{sec:Genre}.} Looking at overall correlation between the model's and professionals' CI, we see a weak and insignificant correlation ($\rho$ = .161, p =.422). Looking at the translation modalities, an interesting difference arises: the correlations almost flatten out for PE and HT, but the correlation increases for MT ($\rho$ = .43) although this is not significant (p = .25). Still, it suggests that the model aligns better with professionals on MT. %translation.
Besides correlation, we can also look at distinguishing texts with CI: for professionals we know see) that CI differentiates modality rather cleanly.\footnote{MT scores lowest, followed by PE and topped by HT (as shown in the figure by the green dots towards the lower end of the Y-axis, the blue in between and the red at the top).} This pattern is %completely
absent from the LLM-as-a-judge CI on the X-axis which shows that CI of the model 
does not %help to 
differentiate between the modalities. We see overall that CI scores from LLM-as-a-judge do not align with professional CI scores and are uninformative for creativity estimation.

%So, LLM-based metrics struggle to annotate literary translation correctly. It has a strong bias towards MT translations and a bias against human translation. This seems to be due in part to the issues it has with marking creative solutions, Kudos, cultural shifts and conventional language use as errors. %It tends to mark these features as errors whereas they are considered to improve literary translation. 
%This is problematic as these features are fundamental for literary translation. If these models are unable to capture these or differentiate between such creative shifts and errors it makes them unsuitable and untrustworthy for annotating literary translation. \rik{Is this meant as a summary paragraph?}

\subsection{%RQ3:
Genre}  
\label{sec:Genre}
Our last RQ investigates whether the correlations between AEMs and LLM-as-a-judge for the first two RQs differ significantly between genres.

\begin{figure}[h]
    \centering
    \includegraphics[scale=0.3]{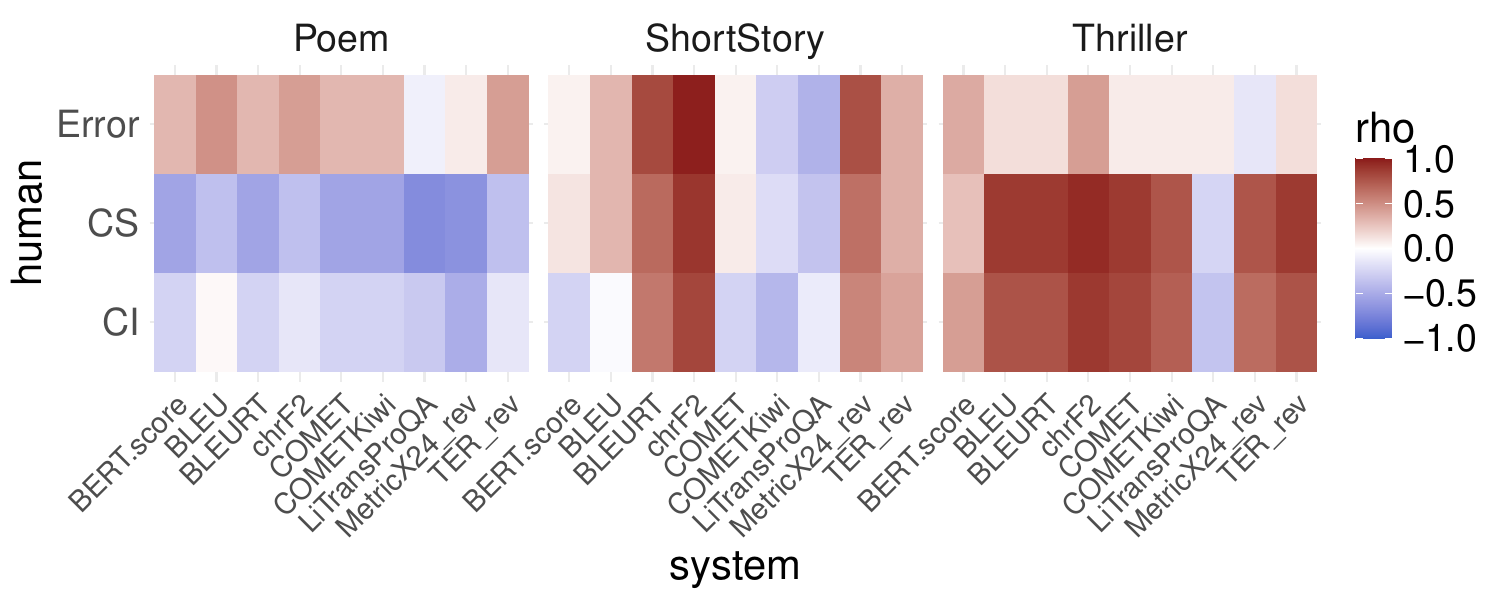}
    \caption{Heat map of Spearman correlations between AEMs and professional evaluations across genres.}
    \label{fig:genreheat}
\end{figure}

%\paragraph{AEMs}
First, we wanted to see if correlations between AEMs and professional annotations were higher for certain genres than for others. To analyse this, we created the same heat map as in Figure \ref{fig:heatmap}, %a heat map of the correlations split
but divided by genre, shown in Figure \ref{fig:genreheat}. Here, we see something striking: correlations between the AEMs and professionals for CSs and CI are much higher for thrillers, especially compared with poems. Our AEM analysis (see Section \ref{sec:results_AEMs}) showed that AEMs struggle with CSs and ignore or even penalise it. Section \ref{sec:hum_ann} further discussed that thrillers are a relatively low and less creative genre compared to high literary genres like poems, which the professional annotations showed as thrillers had significantly fewer CSs and a lower CI than the poems. Perhaps AEMs perform better on thrillers because thrillers are less creative and AEMs perform better on less creative genres. This might also explain why some studies do find higher correlations between AEMs and texts that are less literary, such as fan fiction, as used in WMT25 \cite{kocmi-etal-2025-findings}. It will be interesting to test this on a larger corpus to see if the trend remains.

% but the difference in model performance is stark, with correlations between AEMs and CS significantly higher for thriller compared to poems for all AEMs except BERTscore and LiTransProQA. Above\rik{What does this refer to}, we found t Perhaps it can also help us understand this somewhat conflicting finding here. Thrillers are considered less literary and contain fewer CSs, so AEMs might perform better on less creative genres, compared to highly creative ones like poems. \rik{So why do we call it surprising above?} Potentially this explains why some studies do find higher correlations between AEMs and texts that are less literary, such as fan fiction, as used in WMT25 \cite{kocmi-etal-2025-findings}. It will be interesting to test this on a larger corpus to see if the trend remains.

%\paragraph{LLM-as-a-judge} 
%LLM-as-a-judge also performs poorly of errors and we wanted to see if this differed across genres as well.

\begin{table}[h!]
 %\scriptsize
    \centering
        \resizebox{0.8\columnwidth}{!}{
    \setlength{\tabcolsep}{4pt}
    \begin{tabular}{c |c c c }  
    \toprule
    \textbf{\# Errors} & \textbf{Poem} & \textbf{Short Story} & \textbf{Thriller} \\ \midrule
     Human & 84 & 93 & 107 \\
    LLM & 137 & 95 & 93 \\ 
    Match & 42 & 22 & 28 \\
    \bottomrule
    \end{tabular}
    }
     \captionsetup{justification = centering}
    \caption{Number of errors from professionals and LLM-as-a-judge (including matches) across genre. }
    \label{tab:Genre_errors}
\end{table}

We also analysed LLM-as-a-judge's performance on errors across genre, shown in Table \ref{tab:Genre_errors}. We see that while professionals mark most errors in thrillers, followed by short stories and poems, this is the opposite for LLM-as-a-judge. %To see whether this difference is significant, 
%To analyse how LLM-as-a-judge performs with regards to errors across genres, we included Genre as predictor in the binomial GLM comparing LLM and professional error counts (see Section \ref{sec:LLM-as-a-judge}). This showed that across modalities LLM-as-a-judge indeed marks more errors in the poems and short stories compared to thrillers (z = 3.31, p =  .001), and in poems more than in short stories (z = 2.55, p = .011) compared to the professional annotator. %across all modalities. 
%\AGA{Let me give you an example just with this sentence. When you say marks more errors do you refer in all modalities? what is high literature and lower literature in your dataset, why not talk about the three that you have? But do professional annotators also mark more errors in the text or poem? There are so many assumptions here, and the reader has so many questions. The text should show a clear path here} 
This suggests that LLM-as-a-judge struggles with high literary and more creative texts, ranking those lower than professionals would.%\AGA{Does it? Because where is the table with the errors annotated per genre?}

For creativity evaluation by LLM-as-a-judge, genres were included as shapes in Figure \ref{fig:CIscatter}. No pattern arises directly from the genre. The two severe outliers are poems (the RU-NL PE Poem, scoring a CI of 41 from professionals but -53 from the LLM, and the RU-NL MT Poem, scoring a CI of -150 from the professionals and -27 from the LLM). %\AGA{Is this in a table where we can see the results like with professionals? This is in fact what is interesting. These differences. }
This suggests LLM-as-a-judge struggles more with evaluating poetry compared to the other genres. %, as we saw before as well. 
Looking at the distribution along the X-axis (CI\_LLM), we also see that thrillers have received relatively high scores from the model, which is not always matched by the professionals (as seen in the spread along the Y-axis).\footnote{See Appendix \ref{sec:CI-scores} for detailed CI scores.}
%This could indicate even further that the LLM biases genres that are more creative or 'high literature'.  

\section{Conclusion}
%\todo{KG: I've significantly curtailed/shortened the conclusion now; the older version is still with percentage-marks below, but I thought maybe more to the point and not per RQ saves space and gives a clear overview.}
This paper set out to explore automatic evaluation performance on literary translation, specifically focusing on how well it correlates with human evaluation on creativity. 
Our analyses show that both AEMs and LLM-as-a-judge only partially and weakly correlate with professional annotations. They even penalise some creative shifts, such as correctly adapted cultural references.
A closer look at their performances across genres reveals that they have lower correlations with professional annotations for high literary genres like poems compared to 
%a lower and 
a relatively less creative genre like thrillers. %This suggests that automatic metrics struggle and potentially have a bias against texts and genres that are more creative or 'high' literature. 

Although our dataset is small, these results suggest a key problem in the current metrics' ability to assess literary translation, as creative features are exactly what makes a literary translation stand out. This study begins to help explain the discrepancy between what professional annotators see in a literary text as creative and what AEMs or LLMs consider a quality translation, as in WMT2025 \cite{kocmi-etal-2025-findings}. Future work across a larger dataset using more annotators will show further details and differences between professional and automatic evaluation.
%\rik{I think we need to reverse the order here. First we say: although our data set is small and could be improved by X, Y, Z, our results suggest a key problem with ...} 
Publishers who use these metrics to assess MT-quality and PE-rates should consider whether these metrics can accurately give an indication of the creative work needed. We think future work combining both creative shift and error annotations through multi-agent frameworks could reveal more to us about the workings, problems and potential solutions for machine evaluation. We need to find a new way of measuring that reflects creativity in literary translation better, which we believe real interdisciplinary research can achieve.%\todo{AGA: I like this ending but the use of creative shifts and errors is a bit lost}

\section{Funding \& acknowledgments}
This project has received funding from the EU ERC Consolidator Grant 101086819. %\todo{KG: Should we also include all annotators? And should this section be called acknowldgments or funding?}

We would like to thank all translators and annotators who contributed to this project.

\section{Sustainability statement}
We did not run any of our experiments on dedicated local hardware, but queried LLMs such as ChatGPT and Claude both through their web interface and the ChatGPT API. To the best of our knowledge, the average CO$_\mathrm{2}$ emissions of ChatGPT and Claude models are not publicly disclosed, so we will estimate our emissions for both the MT creation and the LLM-as-a-judge evaluations.

For the MT creation, we used LLMs through their web interface. In total, we submitted 63 requests to ChatGPT, 43 requests to Claude and 12 to both Google Translate and DeepL. A calculation by a third party estimates that each message sent to ChatGPT produces approximately 4.32g CO$_\mathrm{2}$ \cite{co2em}, while each message sent to Claude produces approximately 3.5g \cite{co2em-claude}. This would mean creating our MT emitted 0.4kg CO$_\mathrm{2}$.

For the LLM-as-a-judge evaluations we submitted 1783 API requests to ChatGPT, leading to the processing of 1,563,995 tokens (combining inputs and outputs). %Asking ChatGPT we obtain
%the range [2.5, 23.75], depending on the electricity source and assuming 50 Wh per query. 
Using the 4.32g CO$_\mathrm{2}$ figure above, our LLM-as-a-judge experiments would have emitted 7.7kg CO$_\mathrm{2}$. 

In total, we estimate that our experiments emitted about 8.1kg CO$_\mathrm{2}$, which is equivalent of driving about 55 km. 

% \bibliography{\confname}

\bibliography{eamt26}

@inproceedings{vanroy-etal-2023-mateo,
    title = "{MATEO}: {MA}chine {T}ranslation {E}valuation {O}nline",
    author = "Vanroy, Bram  and
      Tezcan, Arda  and
      Macken, Lieve",
    booktitle = "Proceedings of the 24th Annual Conference of the European Association for Machine Translation",
    month = jun,
    year = "2023",
    address = "Tampere, Finland",
    publisher = "European Association for Machine Translation",
    url = "https://aclanthology.org/2023.eamt-1.52",
    pages = "499--500",
}

@misc{zhang2020bertscoreevaluatingtextgeneration,
      title={BERTScore: Evaluating Text Generation with BERT}, 
      author={Tianyi Zhang and Varsha Kishore and Felix Wu and Kilian Q. Weinberger and Yoav Artzi},
      year={2020},
      eprint={1904.09675},
      archivePrefix={arXiv},
      primaryClass={cs.CL},
      url={https://arxiv.org/abs/1904.09675}, 
}

@inproceedings{Papineni2002,
	author={Kishore Papineni and Salim Roukos and Todd Ward and Wei-Jing Zhu},
	year={2002},
	title={Bleu: a Method for Automatic Evaluation of Machine Translation},
	booktitle={Proceedings of the 40th Annual Meeting of the Association for Computational Linguistics},
	address={Philadelphia, Pennsylvania, USA},
    doi = {10.3115/1073083.1073135},
	pages={311--318}
}

@inproceedings{snover-etal-2006-study,
    title = "A Study of Translation Edit Rate with Targeted Human Annotation",
    author = "Matthew Snover  and
      Dorr, Bonnie  and
      Schwartz, Rich  and
      Micciulla, Linnea  and
      Makhoul, John",
    booktitle = "Proceedings of the 7th Conference of the Association for Machine Translation in the Americas: Technical Papers",
    month = aug # " 8-12",
    year = "2006",
    address = "Cambridge, Massachusetts, USA",
    publisher = "Association for Machine Translation in the Americas",
    url = "https://aclanthology.org/2006.amta-papers.25/",
    pages = "223--231"
}

@inproceedings{popovic-2016-chrf,
    title = "chr{F} deconstructed: beta parameters and n-gram weights",
    author = "Popovi{\'c}, Maja",
    editor = {Bojar, Ond{\v{r}}ej  and
      Buck, Christian  and
      Chatterjee, Rajen  and
      Federmann, Christian  and
      Guillou, Liane  and
      Haddow, Barry  and
      Huck, Matthias  and
      Yepes, Antonio Jimeno  and
      N{\'e}v{\'e}ol, Aur{\'e}lie  and
      Neves, Mariana  and
      Pecina, Pavel  and
      Popel, Martin  and
      Koehn, Philipp  and
      Monz, Christof  and
      Negri, Matteo  and
      Post, Matt  and
      Specia, Lucia  and
      Verspoor, Karin  and
      Tiedemann, J{\"o}rg  and
      Turchi, Marco},
    booktitle = "Proceedings of the First Conference on Machine Translation: Volume 2, Shared Task Papers",
    month = aug,
    year = "2016",
    address = "Berlin, Germany",
    publisher = "Association for Computational Linguistics",
    url = "https://aclanthology.org/W16-2341/",
    doi = "10.18653/v1/W16-2341",
    pages = "499--504"
}

@inproceedings{rei-etal-2020-comet,
    title = "{COMET}: A Neural Framework for {MT} Evaluation",
    author = "Rei, Ricardo  and
      Stewart, Craig  and
      Farinha, Ana C  and
      Lavie, Alon",
    editor = "Webber, Bonnie  and
      Cohn, Trevor  and
      He, Yulan  and
      Liu, Yang",
    booktitle = "Proceedings of the 2020 Conference on Empirical Methods in Natural Language Processing (EMNLP)",
    month = nov,
    year = "2020",
    address = "Online",
    publisher = "Association for Computational Linguistics",
    url = "https://aclanthology.org/2020.emnlp-main.213/",
    doi = "10.18653/v1/2020.emnlp-main.213",
    pages = "2685--2702"
}

@inproceedings{sellam2020bleurt,
    title = "{BLEURT}: Learning Robust Metrics for Text Generation",
    author = "Sellam, Thibault  and
      Das, Dipanjan  and
      Parikh, Ankur",
    editor = "Jurafsky, Dan  and
      Chai, Joyce  and
      Schluter, Natalie  and
      Tetreault, Joel",
    booktitle = "Proceedings of the 58th Annual Meeting of the Association for Computational Linguistics",
    month = jul,
    year = "2020",
    address = "Online",
    publisher = "Association for Computational Linguistics",
    url = "https://aclanthology.org/2020.acl-main.704/",
    doi = "10.18653/v1/2020.acl-main.704",
    pages = "7881--7892"
}

@inproceedings{zhang-etal-2025-litransproqa,
    title = "{L}i{T}rans{P}ro{QA}: An {LLM}-based Literary Translation Evaluation Metric with Professional Question Answering",
    author = "Zhang, Ran  and
      Zhao, Wei  and
      Macken, Lieve  and
      Eger, Steffen",
    editor = "Christodoulopoulos, Christos  and
      Chakraborty, Tanmoy  and
      Rose, Carolyn  and
      Peng, Violet",
    booktitle = "Proceedings of the 2025 Conference on Empirical Methods in Natural Language Processing",
    month = nov,
    year = "2025",
    address = "Suzhou, China",
    publisher = "Association for Computational Linguistics",
    url = "https://aclanthology.org/2025.emnlp-main.1482/",
    doi = "10.18653/v1/2025.emnlp-main.1482",
    pages = "29099--29121",
    ISBN = "979-8-89176-332-6"
}

@inproceedings{he-2024-prompting,
    title = "Prompting {C}hat{GPT} for Translation: A Comparative Analysis of Translation Brief and Persona Prompts",
    author = "He, Sui",
    editor = "Scarton, Carolina  and
      Prescott, Charlotte  and
      Bayliss, Chris  and
      Oakley, Chris  and
      Wright, Joanna  and
      Wrigley, Stuart  and
      Song, Xingyi  and
      Gow-Smith, Edward  and
      Bawden, Rachel  and
      S{\'a}nchez-Cartagena, V{\'i}ctor M  and
      Cadwell, Patrick  and
      Lapshinova-Koltunski, Ekaterina  and
      Cabarr{\~a}o, Vera  and
      Chatzitheodorou, Konstantinos  and
      Nurminen, Mary  and
      Kanojia, Diptesh  and
      Moniz, Helena",
    booktitle = "Proceedings of the 25th Annual Conference of the European Association for Machine Translation (Volume 1)",
    month = jun,
    year = "2024",
    address = "Sheffield, UK",
    publisher = "European Association for Machine Translation (EAMT)",
    url = "https://aclanthology.org/2024.eamt-1.27/",
    pages = "316--326"
}

@InProceedings{SharmaPoetry,
author="Priya Sharma  and Tanuja Yadav",
editor="Uddin, Mohammad Shorif
and Singh, Dharm
and Ganokratanaa, Thittaporn
and Kumawat, Gaurav",
title="Poetry and Emotions: Investigating the Limitations of {AI} Translation",
booktitle="Proceedings of International Conference on Innovations in Data Science",
year="2026",
publisher="Springer Nature Singapore",
address="Singapore",
pages="401--409",
isbn="978-981-96-6329-3"
}

@inproceedings{chakrabarty-etal-2021,
    title = "Don{'}t Go Far Off: An Empirical Study on Neural Poetry Translation",
    author = "Chakrabarty, Tuhin  and
      Saakyan, Arkadiy  and
      Muresan, Smaranda",
    editor = "Moens, Marie-Francine  and
      Huang, Xuanjing  and
      Specia, Lucia  and
      Yih, Scott Wen-tau",
    booktitle = "Proceedings of the 2021 Conference on Empirical Methods in Natural Language Processing",
    month = "nov",
    year = "2021",
    address = "Online and Punta Cana, Dominican Republic",
    publisher = "Association for Computational Linguistics",
    url = "https://aclanthology.org/2021.emnlp-main.577/",
    doi = "10.18653/v1/2021.emnlp-main.577",
    pages = "7253--7265"
}

@ARTICLE{Mukherjee2025,
  author={Mukherjee, Aniruddha and Hassija, Vikas and Chamola, Vinay and Gupta, Karunesh Kumar},
  journal={IEEE Open Journal of the Computer Society}, 
  title={A Detailed Comparative Analysis of Automatic Neural Metrics for Machine Translation: {BLEURT} \& {BERTS}core}, 
  year={2025},
  volume={6},
  number={},
  pages={658-668},
  doi={10.1109/OJCS.2025.3560333}}

@inproceedings{hanna-bojar-2021-fine,
    title = "A Fine-Grained Analysis of {BERTS}core",
    author = "Hanna, Michael  and
      Bojar, Ond{\v{r}}ej",
    editor = "Barrault, Loic  and
      Bojar, Ondrej  and
      Bougares, Fethi  and
      Chatterjee, Rajen  and
      Costa-jussa, Marta R.  and
      Federmann, Christian  and
      Fishel, Mark  and
      Fraser, Alexander  and
      Freitag, Markus  and
      Graham, Yvette  and
      Grundkiewicz, Roman  and
      Guzman, Paco  and
      Haddow, Barry  and
      Huck, Matthias  and
      Yepes, Antonio Jimeno  and
      Koehn, Philipp  and
      Kocmi, Tom  and
      Martins, Andre  and
      Morishita, Makoto  and
      Monz, Christof",
    booktitle = "Proceedings of the Sixth Conference on Machine Translation",
    month = nov,
    year = "2021",
    address = "Online",
    publisher = "Association for Computational Linguistics",
    url = "https://aclanthology.org/2021.wmt-1.59/",
    pages = "507--517"
}

@inproceedings{moghe-etal-2023-extrinsic,
    title = "Extrinsic Evaluation of Machine Translation Metrics",
    author = "Nikita Moghe  and Tom Sherborne and Mark Steedman and Alexandra Birch",
    editor = "Rogers, Anna  and
      Boyd-Graber, Jordan  and
      Okazaki, Naoaki",
    booktitle = "Proceedings of the 61st Annual Meeting of the Association for Computational Linguistics (Volume 1: Long Papers)",
    month = jul,
    year = "2023",
    address = "Toronto, Canada",
    publisher = "Association for Computational Linguistics",
    url = "https://aclanthology.org/2023.acl-long.730/",
    doi = "10.18653/v1/2023.acl-long.730",
    pages = "13060--13078"
}

@inproceedings{freitag-etal-2023-results,
    title = "Results of {WMT}23 Metrics Shared Task: Metrics Might Be Guilty but References Are Not Innocent",
    author = "Freitag, Markus  and
      Mathur, Nitika  and
      Lo, Chi-kiu  and
      Avramidis, Eleftherios  and
      Rei, Ricardo  and
      Thompson, Brian  and
      Kocmi, Tom  and
      Blain, Frederic  and
      Deutsch, Daniel  and
      Stewart, Craig  and
      Zerva, Chrysoula  and
      Castilho, Sheila  and
      Lavie, Alon  and
      Foster, George",
    editor = "Koehn, Philipp  and
      Haddow, Barry  and
      Kocmi, Tom  and
      Monz, Christof",
    booktitle = "Proceedings of the Eighth Conference on Machine Translation",
    month = dec,
    year = "2023",
    address = "Singapore",
    publisher = "Association for Computational Linguistics",
    url = "https://aclanthology.org/2023.wmt-1.51/",
    doi = "10.18653/v1/2023.wmt-1.51",
    pages = "578--628"
}

@inproceedings{saadany-orasan-2021-bleu,
    title = "{BLEU}, {METEOR}, {BERTS}core: Evaluation of Metrics Performance in Assessing Critical Translation Errors in Sentiment-Oriented Text",
    author = "Saadany, Hadeel  and
      Orasan, Constantin",
    editor = "Mitkov, Ruslan  and
      Sosoni, Vilelmini  and
      Gigu{\`e}re, Julie Christine  and
      Murgolo, Elena  and
      Deysel, Elizabeth",
    booktitle = "Proceedings of the Translation and Interpreting Technology Online Conference",
    month = jul,
    year = "2021",
    address = "Held Online",
    publisher = "INCOMA Ltd.",
    url = "https://aclanthology.org/2021.triton-1.6/",
    pages = "48--56"
}

@inproceedings{kocmi-etal-2025-findings,
    title = "Findings of the {WMT}25 General Machine Translation Shared Task: Time to Stop Evaluating on Easy Test Sets",
    author = "Kocmi, Tom  and
      Artemova, Ekaterina  and
      Avramidis, Eleftherios  and
      Bawden, Rachel  and
      Bojar, Ond{\v{r}}ej  and
      Dranch, Konstantin  and
      Dvorkovich, Anton  and
      Dukanov, Sergey  and
      Fishel, Mark  and
      Freitag, Markus  and
      Gowda, Thamme  and
      Grundkiewicz, Roman  and
      Haddow, Barry  and
      Karpinska, Marzena  and
      Koehn, Philipp  and
      Lakougna, Howard  and
      Lundin, Jessica  and
      Monz, Christof  and
      Murray, Kenton  and
      Nagata, Masaaki  and
      Perrella, Stefano  and
      Proietti, Lorenzo  and
      Popel, Martin  and
      Popovi{\'c}, Maja  and
      Riley, Parker  and
      Shmatova, Mariya  and
      Steingr{\'i}msson, Steinth{\'o}r  and
      Yankovskaya, Lisa  and
      Zouhar, Vil{\'e}m",
    editor = "Haddow, Barry  and
      Kocmi, Tom  and
      Koehn, Philipp  and
      Monz, Christof",
    booktitle = "Proceedings of the Tenth Conference on Machine Translation",
    month = nov,
    year = "2025",
    address = "Suzhou, China",
    publisher = "Association for Computational Linguistics",
    url = "https://aclanthology.org/2025.wmt-1.22/",
    doi = "10.18653/v1/2025.wmt-1.22",
    pages = "355--413",
    ISBN = "979-8-89176-341-8"
}

@inproceedings{freitag-etal-2022-results,
    title = "Results of {WMT}22 Metrics Shared Task: Stop Using {BLEU} {--} Neural Metrics Are Better and More Robust",
    author = "Freitag, Markus  and
      Rei, Ricardo  and
      Mathur, Nitika  and
      Lo, Chi-kiu  and
      Stewart, Craig  and
      Avramidis, Eleftherios  and
      Kocmi, Tom  and
      Foster, George  and
      Lavie, Alon  and
      Martins, Andr{\'e} F. T.",
    editor = {Koehn, Philipp  and
      Barrault, Lo{\"i}c  and
      Bojar, Ond{\v{r}}ej  and
      Bougares, Fethi  and
      Chatterjee, Rajen  and
      Costa-juss{\`a}, Marta R.  and
      Federmann, Christian  and
      Fishel, Mark  and
      Fraser, Alexander  and
      Freitag, Markus  and
      Graham, Yvette  and
      Grundkiewicz, Roman  and
      Guzman, Paco  and
      Haddow, Barry  and
      Huck, Matthias  and
      Jimeno Yepes, Antonio  and
      Kocmi, Tom  and
      Martins, Andr{\'e}  and
      Morishita, Makoto  and
      Monz, Christof  and
      Nagata, Masaaki  and
      Nakazawa, Toshiaki  and
      Negri, Matteo  and
      N{\'e}v{\'e}ol, Aur{\'e}lie  and
      Neves, Mariana  and
      Popel, Martin  and
      Turchi, Marco  and
      Zampieri, Marcos},
    booktitle = "Proceedings of the Seventh Conference on Machine Translation (WMT)",
    month = dec,
    year = "2022",
    address = "Abu Dhabi, United Arab Emirates (Hybrid)",
    publisher = "Association for Computational Linguistics",
    url = "https://aclanthology.org/2022.wmt-1.2/",
    doi = "10.18653/v1/2022.wmt-1.2",
    pages = "46--68"
}

@inproceedings{zouhar-etal-2024-fine,
    title = "Fine-Tuned Machine Translation Metrics Struggle in Unseen Domains",
    author = "Zouhar, Vil{\'e}m  and
      Ding, Shuoyang  and
      Currey, Anna  and
      Badeka, Tatyana  and
      Wang, Jenyuan  and
      Thompson, Brian",
    editor = "Ku, Lun-Wei  and
      Martins, Andre  and
      Srikumar, Vivek",
    booktitle = "Proceedings of the 62nd Annual Meeting of the Association for Computational Linguistics (Volume 2: Short Papers)",
    month = aug,
    year = "2024",
    address = "Bangkok, Thailand",
    publisher = "Association for Computational Linguistics",
    url = "https://aclanthology.org/2024.acl-short.45/",
    doi = "10.18653/v1/2024.acl-short.45",
    pages = "488--500"
}

@inproceedings{amrhein-etal-2022-aces,
    title = "{ACES}: Translation Accuracy Challenge Sets for Evaluating Machine Translation Metrics",
    author = "Amrhein, Chantal  and
      Moghe, Nikita  and
      Guillou, Liane",
    editor = {Koehn, Philipp  and
      Barrault, Lo{\"i}c  and
      Bojar, Ond{\v{r}}ej  and
      Bougares, Fethi  and
      Chatterjee, Rajen  and
      Costa-juss{\`a}, Marta R.  and
      Federmann, Christian  and
      Fishel, Mark  and
      Fraser, Alexander  and
      Freitag, Markus  and
      Graham, Yvette  and
      Grundkiewicz, Roman  and
      Guzman, Paco  and
      Haddow, Barry  and
      Huck, Matthias  and
      Jimeno Yepes, Antonio  and
      Kocmi, Tom  and
      Martins, Andr{\'e}  and
      Morishita, Makoto  and
      Monz, Christof  and
      Nagata, Masaaki  and
      Nakazawa, Toshiaki  and
      Negri, Matteo  and
      N{\'e}v{\'e}ol, Aur{\'e}lie  and
      Neves, Mariana  and
      Popel, Martin  and
      Turchi, Marco  and
      Zampieri, Marcos},
    booktitle = "Proceedings of the Seventh Conference on Machine Translation (WMT)",
    month = dec,
    year = "2022",
    address = "Abu Dhabi, United Arab Emirates (Hybrid)",
    publisher = "Association for Computational Linguistics",
    url = "https://aclanthology.org/2022.wmt-1.44/",
    doi = "10.18653/v1/2022.wmt-1.44",
    pages = "479--513"
}

@inproceedings{wu-etal-2024-evaluating-automatic,
    title = "Evaluating Automatic Metrics with Incremental Machine Translation Systems",
    author = "Wu, Guojun  and
      Cohen, Shay B  and
      Sennrich, Rico",
    editor = "Al-Onaizan, Yaser  and
      Bansal, Mohit  and
      Chen, Yun-Nung",
    booktitle = "Findings of the Association for Computational Linguistics: EMNLP 2024",
    month = nov,
    year = "2024",
    address = "Miami, Florida, USA",
    publisher = "Association for Computational Linguistics",
    url = "https://aclanthology.org/2024.findings-emnlp.169/",
    doi = "10.18653/v1/2024.findings-emnlp.169",
    pages = "2994--3005"
}

@inproceedings{fernandes-etal-2023-devil,
    title = "The Devil Is in the Errors: Leveraging Large Language Models for Fine-grained Machine Translation Evaluation",
    author = "Fernandes, Patrick  and
      Deutsch, Daniel  and
      Finkelstein, Mara  and
      Riley, Parker  and
      Martins, Andr{\'e}  and
      Neubig, Graham  and
      Garg, Ankush  and
      Clark, Jonathan  and
      Freitag, Markus  and
      Firat, Orhan",
    editor = "Koehn, Philipp  and
      Haddow, Barry  and
      Kocmi, Tom  and
      Monz, Christof",
    booktitle = "Proceedings of the Eighth Conference on Machine Translation",
    month = dec,
    year = "2023",
    address = "Singapore",
    publisher = "Association for Computational Linguistics",
    url = "https://aclanthology.org/2023.wmt-1.100/",
    doi = "10.18653/v1/2023.wmt-1.100",
    pages = "1066--1083"
}

@inproceedings{lu-etal-2024-error,
    title = "Error Analysis Prompting Enables Human-Like Translation Evaluation in Large Language Models",
    author = "Lu, Qingyu  and
      Qiu, Baopu  and
      Ding, Liang  and
      Zhang, Kanjian  and
      Kocmi, Tom  and
      Tao, Dacheng",
    editor = "Ku, Lun-Wei  and
      Martins, Andre  and
      Srikumar, Vivek",
    booktitle = "Findings of the Association for Computational Linguistics: ACL 2024",
    month = aug,
    year = "2024",
    address = "Bangkok, Thailand",
    publisher = "Association for Computational Linguistics",
    url = "https://aclanthology.org/2024.findings-acl.520/",
    doi = "10.18653/v1/2024.findings-acl.520",
    pages = "8801--8816"
}

@article{AbdulGhaffar2024,
  author       = {Nehal Ali AbdulGhaffar},
  title        = {Beyond Literal Meaning: Neural Machine Translation Constraints in Translating the Poetic Depth of {A}l-{M}utanabbi’s "{T}ell {M}y {B}eloved" .},
  journal      = {Evolutionary Studies in Imaginative Culture },
  year         = {2024},
  url          = {https://doi.org/10.70082/esiculture.vi.1829},
  pages        = {364--374}
}

@inproceedings{rei-etal-2022-cometkiwi,
    title = "{C}omet{K}iwi: {IST}-Unbabel 2022 Submission for the Quality Estimation Shared Task",
    author = "Rei, Ricardo  and
      Treviso, Marcos  and
      Guerreiro, Nuno M.  and
      Zerva, Chrysoula  and
      Farinha, Ana C  and
      Maroti, Christine  and
      C. de Souza, Jos{\'e} G.  and
      Glushkova, Taisiya  and
      Alves, Duarte  and
      Coheur, Luisa  and
      Lavie, Alon  and
      Martins, Andr{\'e} F. T.",
    editor = {Koehn, Philipp  and
      Barrault, Lo{\"i}c  and
      Bojar, Ond{\v{r}}ej  and
      Bougares, Fethi  and
      Chatterjee, Rajen  and
      Costa-juss{\`a}, Marta R.  and
      Federmann, Christian  and
      Fishel, Mark  and
      Fraser, Alexander  and
      Freitag, Markus  and
      Graham, Yvette  and
      Grundkiewicz, Roman  and
      Guzman, Paco  and
      Haddow, Barry  and
      Huck, Matthias  and
      Jimeno Yepes, Antonio  and
      Kocmi, Tom  and
      Martins, Andr{\'e}  and
      Morishita, Makoto  and
      Monz, Christof  and
      Nagata, Masaaki  and
      Nakazawa, Toshiaki  and
      Negri, Matteo  and
      N{\'e}v{\'e}ol, Aur{\'e}lie  and
      Neves, Mariana  and
      Popel, Martin  and
      Turchi, Marco  and
      Zampieri, Marcos},
    booktitle = "Proceedings of the Seventh Conference on Machine Translation (WMT)",
    month = dec,
    year = "2022",
    address = "Abu Dhabi, United Arab Emirates (Hybrid)",
    publisher = "Association for Computational Linguistics",
    url = "https://aclanthology.org/2022.wmt-1.60/",
    doi = "10.18653/v1/2022.wmt-1.60",
    pages = "634--645"
}

@inproceedings{juraska-etal-2024-metricx,
    title = "{M}etric{X}-24: The {G}oogle Submission to the {WMT} 2024 Metrics Shared Task",
    author = "Juraska, Juraj  and
      Deutsch, Daniel  and
      Finkelstein, Mara  and
      Freitag, Markus",
    editor = "Haddow, Barry  and
      Kocmi, Tom  and
      Koehn, Philipp  and
      Monz, Christof",
    booktitle = "Proceedings of the Ninth Conference on Machine Translation",
    month = nov,
    year = "2024",
    address = "Miami, Florida, USA",
    publisher = "Association for Computational Linguistics",
    url = "https://aclanthology.org/2024.wmt-1.35/",
    doi = "10.18653/v1/2024.wmt-1.35",
    pages = "492--504"
}

@inproceedings{agrawal-etal-2024-automatic-metrics,
    title = "Can Automatic Metrics Assess High-Quality Translations?",
    author = "Agrawal, Sweta  and
      Farinhas, Ant{\'o}nio  and
      Rei, Ricardo  and
      Martins, Andre",
    editor = "Al-Onaizan, Yaser  and
      Bansal, Mohit  and
      Chen, Yun-Nung",
    booktitle = "Proceedings of the 2024 Conference on Empirical Methods in Natural Language Processing",
    month = nov,
    year = "2024",
    address = "Miami, Florida, USA",
    publisher = "Association for Computational Linguistics",
    url = "https://aclanthology.org/2024.emnlp-main.802/",
    doi = "10.18653/v1/2024.emnlp-main.802",
    pages = "14491--14502"
}

@inproceedings{schmidtova-etal-2024-automatic-metrics,
    title = "Automatic Metrics in Natural Language Generation: A survey of Current Evaluation Practices",
    author = "Schmidtova, Patricia  and
      Mahamood, Saad  and
      Balloccu, Simone  and
      Dusek, Ondrej  and
      Gatt, Albert  and
      Gkatzia, Dimitra  and
      Howcroft, David M.  and
      Platek, Ondrej  and
      Sivaprasad, Adarsa",
    editor = "Mahamood, Saad  and
      Minh, Nguyen Le  and
      Ippolito, Daphne",
    booktitle = "Proceedings of the 17th International Natural Language Generation Conference",
    month = sep,
    year = "2024",
    address = "Tokyo, Japan",
    publisher = "Association for Computational Linguistics",
    url = "https://aclanthology.org/2024.inlg-main.44/",
    doi = "10.18653/v1/2024.inlg-main.44",
    pages = "557--583"
}

@inproceedings{thai-etal-2022-exploring,
    title = "Exploring Document-Level Literary Machine Translation with Parallel Paragraphs from World Literature",
    author = "Thai, Katherine  and
      Karpinska, Marzena  and
      Krishna, Kalpesh  and
      Ray, Bill  and
      Inghilleri, Moira  and
      Wieting, John  and
      Iyyer, Mohit",
    editor = "Goldberg, Yoav  and
      Kozareva, Zornitsa  and
      Zhang, Yue",
    booktitle = "Proceedings of the 2022 Conference on Empirical Methods in Natural Language Processing",
    month = dec,
    year = "2022",
    address = "Abu Dhabi, United Arab Emirates",
    publisher = "Association for Computational Linguistics",
    url = "https://aclanthology.org/2022.emnlp-main.672/",
    doi = "10.18653/v1/2022.emnlp-main.672",
    pages = "9882--9902"
}

@inproceedings{feng-etal-2025-mad,
    title = "{M}-{MAD}: Multidimensional Multi-Agent Debate for Advanced Machine Translation Evaluation",
    author = "Feng, Zhaopeng  and
      Su, Jiayuan  and
      Zheng, Jiamei  and
      Ren, Jiahan  and
      Zhang, Yan  and
      Wu, Jian  and
      Wang, Hongwei  and
      Liu, Zuozhu",
    editor = "Che, Wanxiang  and
      Nabende, Joyce  and
      Shutova, Ekaterina  and
      Pilehvar, Mohammad Taher",
    booktitle = "Proceedings of the 63rd Annual Meeting of the Association for Computational Linguistics (Volume 1: Long Papers)",
    month = jul,
    year = "2025",
    address = "Vienna, Austria",
    publisher = "Association for Computational Linguistics",
    url = "https://aclanthology.org/2025.acl-long.351/",
    doi = "10.18653/v1/2025.acl-long.351",
    pages = "7084--7107",
    ISBN = "979-8-89176-251-0"
}

@article{Klemin,
    author = {Jeremy Klemin},
    title = {The Last Frontier of Machine Translation},
    journal = {The Atlantic},
    year = {2024},
    month = {January},
    url = {https://www.theatlantic.com/technology/archive/2024/01/literary-translation-artificial-intelligence/677038/},
    lastchecked = {12 March 2026}
}

@article{Bookseller,
    author = {Emily Warner},
    title = {‘Dumbing down’ or ‘momentous’ opportunity? Industry divided over AI literary translation},
    journal = {The Bookseller},
    year = {2025},
    month = {July},
    url = {https://www.thebookseller.com/news/dumbing-down-or-momentous-opportunity-industry-divided-over-ai-literary-translation},
    lastchecked = {12 March 2026}
}

@inproceedings{chaganty-etal-2018-price,
    title = "The price of debiasing automatic metrics in natural language evaluation",
    author = "Chaganty, Arun  and
      Mussmann, Stephen  and
      Liang, Percy",
    editor = "Gurevych, Iryna  and
      Miyao, Yusuke",
    booktitle = "Proceedings of the 56th Annual Meeting of the Association for Computational Linguistics (Volume 1: Long Papers)",
    month = jul,
    year = "2018",
    address = "Melbourne, Australia",
    publisher = "Association for Computational Linguistics",
    url = "https://aclanthology.org/P18-1060/",
    doi = "10.18653/v1/P18-1060",
    pages = "643--653"
}

@inproceedings{zhang-etal-2025-good,
    title = "How Good Are {LLM}s for Literary Translation, Really? Literary Translation Evaluation with Humans and {LLM}s",
    author = "Zhang, Ran  and
      Zhao, Wei  and
      Eger, Steffen",
    editor = "Chiruzzo, Luis  and
      Ritter, Alan  and
      Wang, Lu",
    booktitle = "Proceedings of the 2025 Conference of the Nations of the Americas Chapter of the Association for Computational Linguistics: Human Language Technologies (Volume 1: Long Papers)",
    month = apr,
    year = "2025",
    address = "Albuquerque, New Mexico",
    publisher = "Association for Computational Linguistics",
    url = "https://aclanthology.org/2025.naacl-long.548/",
    doi = "10.18653/v1/2025.naacl-long.548",
    pages = "10961--10988",
    ISBN = "979-8-89176-189-6"
}

@inproceedings{klie-etal-2018-inception,
    title = "The {INCE}p{TION} Platform: Machine-Assisted and Knowledge-Oriented Interactive Annotation",
    author = "Klie, Jan-Christoph and Bugert, Michael  and Boullosa, Beto and Eckart de Castilho, Richard and Gurevych, Iryna",
    booktitle = "Proceedings of the 27th International Conference on Computational Linguistics: System Demonstrations",
    year = "2018",
    address = "Santa Fe, New Mexico",
    url = "https://www.aclweb.org/anthology/C18-2002",
    pages = "5--9"
}

@inproceedings{matusov-2019-challenges,
    title = "The Challenges of Using Neural Machine Translation for Literature",
    author = "Matusov, Evgeny",
    editor = "Hadley, James  and
      Popovi{\'c}, Maja  and
      Afli, Haithem  and
      Way, Andy",
    booktitle = "Proceedings of the Qualities of Literary Machine Translation",
    month = aug,
    year = "2019",
    address = "Dublin, Ireland",
    publisher = "European Association for Machine Translation",
    url = "https://aclanthology.org/W19-7302/",
    pages = "10--19"
}

@article{CorpasPastor2023,
  title={Machine vs Human Translation of Formal Neologisms in Literature:
Exploring {E}-tools and Creativity in Students},
  author={Laura Noriega-Santi{\'a}{\~n}ez and Gloria Corpas Pastor},
  journal={Revista Tradum{\`a}tica. Tecnologies de la traducci{\'o}},
  year={2023},
  url={https://api.semanticscholar.org/CorpusID:266941889},
  volume ={21},
  pages ={233--264}
}

@Article{CorpasPastor2024,
AUTHOR = {Corpas Pastor, Gloria and Noriega-Santiáñez, Laura},
TITLE = {Human versus Neural Machine Translation Creativity: A Study on Manipulated MWEs in Literature},
JOURNAL = {Information},
VOLUME = {15},
YEAR = {2024},
NUMBER = {9},
ARTICLE-NUMBER = {530},
URL = {https://www.mdpi.com/2078-2489/15/9/530},
ISSN = {2078-2489},
DOI = {10.3390/info15090530}
}

@inproceedings{karpinska-iyyer-2023-large,
    title = "Large Language Models Effectively Leverage Document-level Context for Literary Translation, but Critical Errors Persist",
    author = "Karpinska, Marzena  and
      Iyyer, Mohit",
    editor = "Koehn, Philipp  and
      Haddow, Barry  and
      Kocmi, Tom  and
      Monz, Christof",
    booktitle = "Proceedings of the Eighth Conference on Machine Translation",
    month = dec,
    year = "2023",
    address = "Singapore",
    publisher = "Association for Computational Linguistics",
    url = "https://aclanthology.org/2023.wmt-1.41/",
    doi = "10.18653/v1/2023.wmt-1.41",
    pages = "419--451"
}

@article{KennyWinters,
    author = {Dorothy Kenny and Marion Winters},
    title = {Machine translation, ethics and the literary translator’s voice},
    journal = {Translation Spaces},
    volume ={9},
    number = {1},
    pages = {123--149},
    year = {2020}, 
    doi = {10.1075/ts.00024.ken}
}

@article{Mohar2020,
    author = { Tjaša Mohar and Sara Orthaber and Tomaž Onič},
    title = { Machine Translated {Atwood}: Utopia or Dystopia?},
    journal = { ELOPE: English Language Overseas Perspectives and Enquiries},
    volume = {17},
    number ={1}, 
    pages ={125--141},
    year = {2020},
    doi = {10.4312/elope.17.1.125-141}
}

@article{Guerberof2022,
	Author = {Ana Guerberof-Arenas and Antonio Toral},
	Journal = {Translation Space},
	Title = {Creativity in Translation: Machine Translation as a Constraint for Literary Texts},
	Volume = {11},
    Number ={2},
    Pages = {184--212},
	Year = {2022}
}

@article{Guerberof2020,
	Author = {Ana Guerberof-Arenas and Antonio Toral},
	Journal = {Translation Journal},
	Title = {The Impact of Post-Editing and Machine Translation on Creativity and Reading Experience},
	Volume = {9},
    Number ={2},
    Pages = {255--282},
	Year = {2020}
}

@inproceedings{belouadi-eger-2023-bygpt5,
    title = "{B}y{GPT}5: End-to-End Style-conditioned Poetry Generation with Token-free Language Models",
    author = "Belouadi, Jonas  and
      Eger, Steffen",
    editor = "Rogers, Anna  and
      Boyd-Graber, Jordan  and
      Okazaki, Naoaki",
    booktitle = "Proceedings of the 61st Annual Meeting of the Association for Computational Linguistics (Volume 1: Long Papers)",
    month = jul,
    year = "2023",
    address = "Toronto, Canada",
    publisher = "Association for Computational Linguistics",
    url = "https://aclanthology.org/2023.acl-long.406/",
    doi = "10.18653/v1/2023.acl-long.406",
    pages = "7364--7381"
}

@inproceedings{chen-etal-2024-evaluating-diversity,
    title = "Evaluating Diversity in Automatic Poetry Generation",
    author = {Chen, Yanran  and
      Gr{\"o}ner, Hannes  and
      Zarrie{\ss}, Sina  and
      Eger, Steffen},
    editor = "Al-Onaizan, Yaser  and
      Bansal, Mohit  and
      Chen, Yun-Nung",
    booktitle = "Proceedings of the 2024 Conference on Empirical Methods in Natural Language Processing",
    month = nov,
    year = "2024",
    address = "Miami, Florida, USA",
    publisher = "Association for Computational Linguistics",
    url = "https://aclanthology.org/2024.emnlp-main.1097/",
    doi = "10.18653/v1/2024.emnlp-main.1097",
    pages = "19671--19692"
}

@inproceedings{Chakrabarty2024,
author = {Chakrabarty, Tuhin and Laban, Philippe and Agarwal, Divyansh and Muresan, Smaranda and Wu, Chien-Sheng},
title = {Art or Artifice? Large Language Models and the False Promise of Creativity},
year = {2024},
isbn = {9798400703300},
publisher = {Association for Computing Machinery},
address = {New York, NY, USA},
url = {https://doi.org/10.1145/3613904.3642731},
doi = {10.1145/3613904.3642731},
booktitle = {Proceedings of the 2024 CHI Conference on Human Factors in Computing Systems},
articleno = {30},
numpages = {34},
location = {Honolulu, HI, USA},
series = {CHI '24}
}

@inproceedings{callison-burch-etal-2006-evaluating,
    title = "Re-evaluating the Role of {B}leu in Machine Translation Research",
    author = "Callison-Burch, Chris  and
      Osborne, Miles  and
      Koehn, Philipp",
    editor = "McCarthy, Diana  and
      Wintner, Shuly",
    booktitle = "11th Conference of the {E}uropean Chapter of the Association for Computational Linguistics",
    month = apr,
    year = "2006",
    address = "Trento, Italy",
    publisher = "Association for Computational Linguistics",
    url = "https://aclanthology.org/E06-1032/",
    pages = "249--256"
}

@inproceedings{mathur-etal-2020-tangled,
    title = "Tangled up in {BLEU}: Reevaluating the Evaluation of Automatic Machine Translation Evaluation Metrics",
    author = "Mathur, Nitika  and
      Baldwin, Timothy  and
      Cohn, Trevor",
    editor = "Jurafsky, Dan  and
      Chai, Joyce  and
      Schluter, Natalie  and
      Tetreault, Joel",
    booktitle = "Proceedings of the 58th Annual Meeting of the Association for Computational Linguistics",
    month = jul,
    year = "2020",
    address = "Online",
    publisher = "Association for Computational Linguistics",
    url = "https://aclanthology.org/2020.acl-main.448/",
    doi = "10.18653/v1/2020.acl-main.448",
    pages = "4984--4997"
}

@inproceedings{novikova-etal-2017-need,
    title = "Why We Need New Evaluation Metrics for {NLG}",
    author = "Novikova, Jekaterina  and
      Du{\v{s}}ek, Ond{\v{r}}ej  and
      Cercas Curry, Amanda  and
      Rieser, Verena",
    editor = "Palmer, Martha  and
      Hwa, Rebecca  and
      Riedel, Sebastian",
    booktitle = "Proceedings of the 2017 Conference on Empirical Methods in Natural Language Processing",
    month = sep,
    year = "2017",
    address = "Copenhagen, Denmark",
    publisher = "Association for Computational Linguistics",
    url = "https://aclanthology.org/D17-1238/",
    doi = "10.18653/v1/D17-1238",
    pages = "2241--2252"
}

@inproceedings{zouhar-etal-2025-ai,
    title = "{AI}-Assisted Human Evaluation of Machine Translation",
    author = "Zouhar, Vil{\'e}m  and
      Kocmi, Tom  and
      Sachan, Mrinmaya",
    editor = "Chiruzzo, Luis  and
      Ritter, Alan  and
      Wang, Lu",
    booktitle = "Proceedings of the 2025 Conference of the Nations of the Americas Chapter of the Association for Computational Linguistics: Human Language Technologies (Volume 1: Long Papers)",
    month = apr,
    year = "2025",
    address = "Albuquerque, New Mexico",
    publisher = "Association for Computational Linguistics",
    url = "https://aclanthology.org/2025.naacl-long.255/",
    doi = "10.18653/v1/2025.naacl-long.255",
    pages = "4936--4950",
    ISBN = "979-8-89176-189-6"
}

@inproceedings{xu-etal-2023-instructscore,
    title = "{INSTRUCTSCORE}: Towards Explainable Text Generation Evaluation with Automatic Feedback",
    author = "Xu, Wenda  and
      Wang, Danqing  and
      Pan, Liangming  and
      Song, Zhenqiao  and
      Freitag, Markus  and
      Wang, William  and
      Li, Lei",
    editor = "Bouamor, Houda  and
      Pino, Juan  and
      Bali, Kalika",
    booktitle = "Proceedings of the 2023 Conference on Empirical Methods in Natural Language Processing",
    month = dec,
    year = "2023",
    address = "Singapore",
    publisher = "Association for Computational Linguistics",
    url = "https://aclanthology.org/2023.emnlp-main.365/",
    doi = "10.18653/v1/2023.emnlp-main.365",
    pages = "5967--5994"
}

@inproceedings{liu-etal-2023-g,
    title = "{G}-Eval: {NLG} Evaluation using Gpt-4 with Better Human Alignment",
    author = "Liu, Yang  and
      Iter, Dan  and
      Xu, Yichong  and
      Wang, Shuohang  and
      Xu, Ruochen  and
      Zhu, Chenguang",
    editor = "Bouamor, Houda  and
      Pino, Juan  and
      Bali, Kalika",
    booktitle = "Proceedings of the 2023 Conference on Empirical Methods in Natural Language Processing",
    month = dec,
    year = "2023",
    address = "Singapore",
    publisher = "Association for Computational Linguistics",
    url = "https://aclanthology.org/2023.emnlp-main.153/",
    doi = "10.18653/v1/2023.emnlp-main.153",
    pages = "2511--2522"
}

@inproceedings{yeom-etal-2025-tagged,
    title = "Tagged Span Annotation for Detecting Translation Errors in Reasoning {LLM}s",
    author = "Yeom, Taemin  and
      Ryu, Yonghyun  and
      Choi, Yoonjung  and
      Bak, Jinyeong",
    editor = "Haddow, Barry  and
      Kocmi, Tom  and
      Koehn, Philipp  and
      Monz, Christof",
    booktitle = "Proceedings of the Tenth Conference on Machine Translation",
    month = nov,
    year = "2025",
    address = "Suzhou, China",
    publisher = "Association for Computational Linguistics",
    url = "https://aclanthology.org/2025.wmt-1.62/",
    doi = "10.18653/v1/2025.wmt-1.62",
    pages = "878--886",
    ISBN = "979-8-89176-341-8"
}

@inproceedings{qiu-hu-2025-deep,
    title = "Deep Associations, High Creativity: A Simple yet Effective Metric for Evaluating Large Language Models",
    author = "Qiu, Ziliang  and
      Hu, Renfen",
    editor = "Christodoulopoulos, Christos  and
      Chakraborty, Tanmoy  and
      Rose, Carolyn  and
      Peng, Violet",
    booktitle = "Proceedings of the 2025 Conference on Empirical Methods in Natural Language Processing",
    month = nov,
    year = "2025",
    address = "Suzhou, China",
    publisher = "Association for Computational Linguistics",
    url = "https://aclanthology.org/2025.emnlp-main.550/",
    doi = "10.18653/v1/2025.emnlp-main.550",
    pages = "10859--10872",
    ISBN = "979-8-89176-332-6"
}

@article{Bielova_2025, 
title={Machine vs. Human Translation of Stylistic Neologisms in English Language Chick Lit into Ukrainian}, url={https://www.journals.vu.lt/respectus-philologicus/article/view/38824}, DOI={10.15388/RESPECTUS.2025.48.9}, abstractNote={
Stylistic neologisms (SN), new words crafted to achieve pragmatic effects, pose significant challenges for translators, particularly between typologically distant languages like English and Ukrainian. With advances in machine translation (MT), evaluating its handling of SNs – especially vs. human translation (HT) – is crucial. Despite growing research on HT vs MT of neologisms, studies focusing on English > Ukrainian translation remain absent, leaving a critical research gap. This study addresses this gap by analysing how HT and MT rendered SNs formed through various morphological patterns in English language chick (ELCL) into Ukrainian. The findings reveal that while HT demonstrated exceptional abilities to creatively render English language SNs into Ukrainian, MT lacked creative and cultural sensitivity. MT exhibited an error rate of over 80%, significantly higher than HT’s 14%, with the most frequent errors occurring with loss of connotative meaning and incorrect word formation. The lack of innovation and contextual awareness in MT outputs underscores the necessity for algorithmic advancements and post-editing strategies to improve the rendering of SNs in literary translation.
}, number={48 (53)}, journal={Respectus Philologicus}, author={Bielova, Maryna}, year={2025}, month={Oct.}, pages={109–123} }

@misc{atmakuru2024cs4measuringcreativitylarge,
      title={CS4: Measuring the Creativity of Large Language Models Automatically by Controlling the Number of Story-Writing Constraints}, 
      author={Anirudh Atmakuru and Jatin Nainani and Rohith Siddhartha Reddy Bheemreddy and Anirudh Lakkaraju and Zonghai Yao and Hamed Zamani and Haw-Shiuan Chang},
      year={2024},
      eprint={2410.04197},
      archivePrefix={arXiv},
      primaryClass={cs.CL},
      url={https://arxiv.org/abs/2410.04197}, 
}

@ARTICLE{Kovalkov2021,
  author={Kovalkov, Anastasia and Paaßen, Benjamin and Segal, Avi and Pinkwart, Niels and Gal, Kobi},
  journal={IEEE Transactions on Learning Technologies}, 
  title={Automatic Creativity Measurement in Scratch Programs Across Modalities}, 
  year={2021},
  volume={14},
  number={6},
  pages={740-753},
  doi={10.1109/TLT.2022.3144442}}

@inproceedings{castaldo-monti-2024-prompting,
    title = "Prompting Large Language Models for Idiomatic Translation",
    author = "Castaldo, Antonio  and
      Monti, Johanna",
    editor = "Vanroy, Bram  and
      Lefer, Marie-Aude  and
      Macken, Lieve  and
      Ruffo, Paola",
    booktitle = "Proceedings of the 1st Workshop on Creative-text Translation and Technology",
    month = jun,
    year = "2024",
    address = "Sheffield, United Kingdom",
    publisher = "European Association for Machine Translation",
    url = "https://aclanthology.org/2024.ctt-1.4/",
    pages = "32--39"
}

@inproceedings{zhang-etal-2023-machine,
    title = "Machine Translation with Large Language Models: Prompting, Few-shot Learning, and Fine-tuning with {QL}o{RA}",
    author = "Zhang, Xuan  and
      Rajabi, Navid  and
      Duh, Kevin  and
      Koehn, Philipp",
    editor = "Koehn, Philipp  and
      Haddow, Barry  and
      Kocmi, Tom  and
      Monz, Christof",
    booktitle = "Proceedings of the Eighth Conference on Machine Translation",
    month = dec,
    year = "2023",
    address = "Singapore",
    publisher = "Association for Computational Linguistics",
    url = "https://aclanthology.org/2023.wmt-1.43/",
    doi = "10.18653/v1/2023.wmt-1.43",
    pages = "468--481"
}

@InProceedings{pmlr-v202-zhang23m,
  title = 	 {Prompting Large Language Model for Machine Translation: A Case Study},
  author =       {Zhang, Biao and Haddow, Barry and Birch, Alexandra},
  booktitle = 	 {Proceedings of the 40th International Conference on Machine Learning},
  pages = 	 {41092--41110},
  year = 	 {2023},
  editor = 	 {Krause, Andreas and Brunskill, Emma and Cho, Kyunghyun and Engelhardt, Barbara and Sabato, Sivan and Scarlett, Jonathan},
  volume = 	 {202},
  series = 	 {Proceedings of Machine Learning Research},
  month = 	 {23--29 Jul},
  publisher =    {PMLR},
  url = 	 {https://proceedings.mlr.press/v202/zhang23m.html}
}

@inproceedings{Gao2024,
author = {Gao, Yuan and Wang, Ruili and Hou, Feng},
title = {How to Design Translation Prompts for ChatGPT: An Empirical Study},
year = {2024},
isbn = {9798400713149},
publisher = {Association for Computing Machinery},
address = {New York, NY, USA},
url = {https://doi.org/10.1145/3700410.3702123},
doi = {10.1145/3700410.3702123},
booktitle = {Proceedings of the 6th ACM International Conference on Multimedia in Asia Workshops},
articleno = {9},
numpages = {7},
location = {
},
series = {MMAsia '24 Workshops}
}

@inproceedings{yamada-2023-optimizing,
    title = "Optimizing Machine Translation through Prompt Engineering: An Investigation into {C}hat{GPT}{'}s Customizability",
    author = "Yamada, Masaru",
    editor = "Yamada, Masaru  and
      do Carmo, Felix",
    booktitle = "Proceedings of Machine Translation Summit XIX, Vol. 2: Users Track",
    month = sep,
    year = "2023",
    address = "Macau SAR, China",
    publisher = "Asia-Pacific Association for Machine Translation",
    url = "https://aclanthology.org/2023.mtsummit-users.19/",
    pages = "195--204"
}

@article{Opaluwah_2025,
   title={Prompt-oriented Output of Culture-Specific Items in Translated African Poetry by Large Language Models: An Initial Multi-layered Tabular Review},
   volume={8},
   ISSN={2640-0111},
   url={http://dx.doi.org/10.11648/j.ajcst.20250802.14},
   DOI={10.11648/j.ajcst.20250802.14},
   number={2},
   journal={American Journal of Computer Science and Technology},
   publisher={Science Publishing Group},
   author={Opaluwah, Adeyola},
   year={2025},
   month=jun, pages={85–101} }

@inproceedings{cui-etal-2024-efficiently,
    title = "Efficiently Exploring Large Language Models for Document-Level Machine Translation with In-context Learning",
    author = "Cui, Menglong  and
      Du, Jiangcun  and
      Zhu, Shaolin  and
      Xiong, Deyi",
    editor = "Ku, Lun-Wei  and
      Martins, Andre  and
      Srikumar, Vivek",
    booktitle = "Findings of the Association for Computational Linguistics: ACL 2024",
    month = aug,
    year = "2024",
    address = "Bangkok, Thailand",
    publisher = "Association for Computational Linguistics",
    url = "https://aclanthology.org/2024.findings-acl.646/",
    doi = "10.18653/v1/2024.findings-acl.646",
    pages = "10885--10897"
}

@inproceedings{egdom-etal-2024-make,
    title = "`Can make mistakes'. Prompting {C}hat{GPT} to Enhance Literary {MT} output",
    author = "Egdom, Gys-Walt  and
      Declercq, Christophe  and
      Kosters, Onno",
    editor = "Vanroy, Bram  and
      Lefer, Marie-Aude  and
      Macken, Lieve  and
      Ruffo, Paola",
    booktitle = "Proceedings of the 1st Workshop on Creative-text Translation and Technology",
    month = jun,
    year = "2024",
    address = "Sheffield, United Kingdom",
    publisher = "European Association for Machine Translation",
    url = "https://aclanthology.org/2024.ctt-1.2/",
    pages = "10--20"
}

@article{Bayer-Hohenwarter2011,
	Author = {Gerrit Bayer-Hohenwarter},
	Journal = {Meta},
	Pages = {663--692},
	Title = {Creative shifts as a means of measuring and promoting translation Creativity},
	Volume = {56},
    Number = {3},
	Year = {2011},
    doi = {10.7202/1008339ar}
}

@inproceedings{castaldo-etal-2025-extending,
    title = "Extending {CREAMT}: Leveraging Large Language Models for Literary Translation Post-Editing",
    author = "Castaldo, Antonio  and
      Castilho, Sheila  and
      Moorkens, Joss  and
      Monti, Johanna",
    editor = "Bouillon, Pierrette  and
      Gerlach, Johanna  and
      Girletti, Sabrina  and
      Volkart, Lise  and
      Rubino, Raphael  and
      Sennrich, Rico  and
      Farinha, Ana C.  and
      Gaido, Marco  and
      Daems, Joke  and
      Kenny, Dorothy  and
      Moniz, Helena  and
      Szoc, Sara",
    booktitle = "Proceedings of Machine Translation Summit XX: Volume 1",
    month = jun,
    year = "2025",
    address = "Geneva, Switzerland",
    publisher = "European Association for Machine Translation",
    url = "https://aclanthology.org/2025.mtsummit-1.40/",
    pages = "506--515",
    ISBN = "978-2-9701897-0-1"
}

@inproceedings{gerrits-arenas-2025-mt,
    title = "To {MT} or not to {MT}: An eye-tracking study on the reception by {D}utch readers of different translation and creativity levels",
    author = "Gerrits, Kyo  and
      Guerberof-Arenas, Ana",
    editor = "Bouillon, Pierrette  and
      Gerlach, Johanna  and
      Girletti, Sabrina  and
      Volkart, Lise  and
      Rubino, Raphael  and
      Sennrich, Rico  and
      Farinha, Ana C.  and
      Gaido, Marco  and
      Daems, Joke  and
      Kenny, Dorothy  and
      Moniz, Helena  and
      Szoc, Sara",
    booktitle = "Proceedings of Machine Translation Summit XX: Volume 1",
    month = jun,
    year = "2025",
    address = "Geneva, Switzerland",
    publisher = "European Association for Machine Translation",
    url = "https://aclanthology.org/2025.mtsummit-1.41/",
    pages = "516--537",
    ISBN = "978-2-9701897-0-1"
}

@article{Tewari2026,
	Author = {Pragya Tewari and Anurag Singh Baghel},
	Journal = {Journal of Theoretical and Applied Information Technology},
	Pages = {560--569},
	Title = {STYLISTICALLY-AWARE {H}INDI-{E}NGLISH POETIC TRANSLATION WITH MBART AND {LLM}-BASED POST-EDITING},
	Volume = {104},
    Number = {3},
	Year = {2026},
    doi = {10.5281/zenodo.18667066}
}

@inproceedings{daems-etal-2024-impact,
    title = "Impact of translation workflows with and without {MT} on textual characteristics in literary translation",
    author = "Daems, Joke  and
      Ruffo, Paola  and
      Macken, Lieve",
    editor = "Vanroy, Bram  and
      Lefer, Marie-Aude  and
      Macken, Lieve  and
      Ruffo, Paola",
    booktitle = "Proceedings of the 1st Workshop on Creative-text Translation and Technology",
    month = jun,
    year = "2024",
    address = "Sheffield, United Kingdom",
    publisher = "European Association for Machine Translation",
    url = "https://aclanthology.org/2024.ctt-1.6/",
    pages = "57--64"
}

@inproceedings{macken-etal-2022-literary,
    title = "Literary translation as a three-stage process: machine translation, post-editing and revision",
    author = "Macken, Lieve  and
      Vanroy, Bram  and
      Desmet, Luca  and
      Tezcan, Arda",
    editor = {Moniz, Helena  and
      Macken, Lieve  and
      Rufener, Andrew  and
      Barrault, Lo{\"i}c  and
      Costa-juss{\`a}, Marta R.  and
      Declercq, Christophe  and
      Koponen, Maarit  and
      Kemp, Ellie  and
      Pilos, Spyridon  and
      Forcada, Mikel L.  and
      Scarton, Carolina  and
      Van den Bogaert, Joachim  and
      Daems, Joke  and
      Tezcan, Arda  and
      Vanroy, Bram  and
      Fonteyne, Margot},
    booktitle = "Proceedings of the 23rd Annual Conference of the European Association for Machine Translation",
    month = jun,
    year = "2022",
    address = "Ghent, Belgium",
    publisher = "European Association for Machine Translation",
    url = "https://aclanthology.org/2022.eamt-1.13/",
    pages = "101--110"
}

@misc{kim2025maslitevalmultiagentliterary,
      title={MAS-LitEval : Multi-Agent System for Literary Translation Quality Assessment}, 
      author={Junghwan Kim and Kieun Park and Sohee Park and Hyunggug Kim and Bongwon Suh},
      year={2025},
      eprint={2506.14199},
      archivePrefix={arXiv},
      primaryClass={cs.CL},
      url={https://arxiv.org/abs/2506.14199}, 
}

@incollection{Kussmaul1991,
    author = {Paul Kussmaul},
    title = {Creativity in the Translation Process: Empirical Approaches},
    booktitle = {Translation Studies: The State of the Art. Proceedings of the First James S. Holmes Symposium on Translation Studies},
    publisher = {Rodopi},
    pages = {91--101},
    year = {1991},
    editor = {Kitty M. Leuven-Zwart
and Ton Naaijkens},
    address = {Amsterdam, NL},
    doi = {10.1163/9789004488106_011}
}

@book{Kussmaul1995,
    author = {Paul Kussmaul},
    title = {Training the Translator},
    publisher = {John Benjamins Publishing},
    year = {1995},
    address = {Amsterdam, NL},
    doi = {10.1075/btl.10}
}

@incollection{Kussmaul2000a,
    author = {Paul Kussmaul},
    title = {A Cognitive Framework for Looking at Creative Mental Processes},
    booktitle = {Intercultural Faultlines Research Models in Translation Studies: v. 1: Textual and Cognitive Aspects},
    editor ={Maeve Olohan},
    publisher = {Routledge},
    address = {London, UK},
    year = {2000},
    pages ={59--71},
    doi = {10.4324/9781315759951-5}
}

@incollection{Bayer-Hohenwarter2009,
    author = {Gerrit Bayer-Hohenwarter},
    title = {Translation Creativity: How to Measure the Unmeasurable},
    booktitle = {Behind the Mind: Methods, Models and Results in Translation Process Research},
    editor = {Susanne Göpferich and Arnt Lykke Jakobsen and Inger M. Mees},
    publisher = {Samfundslitteratur},
    year = {2009},
    pages = {39--59},
    address = {Copenhagen, DK}
}

@inproceedings{Bayer-Hohenwarter2013,
    author = {Gerrit Bayer-Hohenwarter},
    title = {Triangulating Translation Creativity Scores} ,
    booktitle = {Tracks and Treks in Translation Studies: Selected Papers from the EST Conference, Leuven 2010},
    pages = {63--85},        
    year = {2013},
    doi ={doi.org/10.1075/btl.108.04bay}
}

@article{Kaufman01012012,
author = {James C. Kaufman and John Baer},
title = {Beyond New and Appropriate: Who Decides What Is Creative?},
journal = {Creativity Research Journal},
volume = {24},
number = {1},
pages = {83--91},
year = {2012},
publisher = {Routledge},
doi = {10.1080/10400419.2012.649237},


URL = { 
    
        https://doi.org/10.1080/10400419.2012.649237
    
    

},
eprint = { 
    
        https://doi.org/10.1080/10400419.2012.649237
    
    

}

}

@inproceedings{kocmi-federmann-2023-gemba,
    title = "{GEMBA}-{MQM}: Detecting Translation Quality Error Spans with {GPT}-4",
    author = "Kocmi, Tom  and
      Federmann, Christian",
    editor = "Koehn, Philipp  and
      Haddow, Barry  and
      Kocmi, Tom  and
      Monz, Christof",
    booktitle = "Proceedings of the Eighth Conference on Machine Translation",
    month = dec,
    year = "2023",
    address = "Singapore",
    publisher = "Association for Computational Linguistics",
    url = "https://aclanthology.org/2023.wmt-1.64/",
    doi = "10.18653/v1/2023.wmt-1.64",
    pages = "768--775"
}

@article{creamer_dutch_2024,
	chapter = {Books},
	title = {Dutch publisher to use {AI} to translate ‘limited number of books’ into {English}},
	issn = {0261-3077},
	url = {https://www.theguardian.com/books/2024/nov/04/dutch-publisher-to-use-ai-to-translate-books-into-english-veen-bosch-keuning-artificial-intelligence},
	language = {en-GB},
	urldate = {2024-12-10},
	journal = {The Guardian},
	author = {Creamer, Ella},
	month = nov,
	year = {2024},
}

@inproceedings{fu-etal-2024-gptscore,
    title = "{GPTS}core: Evaluate as You Desire",
    author = "Fu, Jinlan  and
      Ng, See-Kiong  and
      Jiang, Zhengbao  and
      Liu, Pengfei",
    editor = "Duh, Kevin  and
      Gomez, Helena  and
      Bethard, Steven",
    booktitle = "Proceedings of the 2024 Conference of the North American Chapter of the Association for Computational Linguistics: Human Language Technologies (Volume 1: Long Papers)",
    month = jun,
    year = "2024",
    address = "Mexico City, Mexico",
    publisher = "Association for Computational Linguistics",
    url = "https://aclanthology.org/2024.naacl-long.365/",
    doi = "10.18653/v1/2024.naacl-long.365",
    pages = "6556--6576"
}

@inproceedings{wang-etal-2023-chatgpt,
    title = "Is {C}hat{GPT} a Good {NLG} Evaluator? A Preliminary Study",
    author = "Wang, Jiaan  and
      Liang, Yunlong  and
      Meng, Fandong  and
      Sun, Zengkui  and
      Shi, Haoxiang  and
      Li, Zhixu  and
      Xu, Jinan  and
      Qu, Jianfeng  and
      Zhou, Jie",
    editor = "Dong, Yue  and
      Xiao, Wen  and
      Wang, Lu  and
      Liu, Fei  and
      Carenini, Giuseppe",
    booktitle = "Proceedings of the 4th New Frontiers in Summarization Workshop",
    month = dec,
    year = "2023",
    address = "Singapore",
    publisher = "Association for Computational Linguistics",
    url = "https://aclanthology.org/2023.newsum-1.1/",
    doi = "10.18653/v1/2023.newsum-1.1",
    pages = "1--11"
}

@inproceedings{zhang-etal-2025-crowd,
    title = "Crowd Comparative Reasoning: Unlocking Comprehensive Evaluations for {LLM}-as-a-Judge",
    author = "Zhang, Qiyuan  and
      Wang, Yufei  and
      Jiang, Yuxin  and
      Li, Liangyou  and
      Wu, Chuhan  and
      Wang, Yasheng  and
      Jiang, Xin  and
      Shang, Lifeng  and
      Tang, Ruiming  and
      Lyu, Fuyuan  and
      Ma, Chen",
    editor = "Che, Wanxiang  and
      Nabende, Joyce  and
      Shutova, Ekaterina  and
      Pilehvar, Mohammad Taher",
    booktitle = "Proceedings of the 63rd Annual Meeting of the Association for Computational Linguistics (Volume 1: Long Papers)",
    month = jul,
    year = "2025",
    address = "Vienna, Austria",
    publisher = "Association for Computational Linguistics",
    url = "https://aclanthology.org/2025.acl-long.252/",
    doi = "10.18653/v1/2025.acl-long.252",
    pages = "5059--5074",
    ISBN = "979-8-89176-251-0"
}

@inproceedings{lavie-etal-2025-findings,
    title = "Findings of the {WMT}25 Shared Task on Automated Translation Evaluation Systems: Linguistic Diversity is Challenging and References Still Help",
    author = "Lavie, Alon  and
      Hanneman, Greg  and
      Agrawal, Sweta  and
      Kanojia, Diptesh  and
      Lo, Chi-Kiu  and
      Zouhar, Vil{\'e}m  and
      Blain, Frederic  and
      Zerva, Chrysoula  and
      Avramidis, Eleftherios  and
      Deoghare, Sourabh  and
      Sindhujan, Archchana  and
      Wang, Jiayi  and
      Adelani, David Ifeoluwa  and
      Thompson, Brian  and
      Kocmi, Tom  and
      Freitag, Markus  and
      Deutsch, Daniel",
    editor = "Haddow, Barry  and
      Kocmi, Tom  and
      Koehn, Philipp  and
      Monz, Christof",
    booktitle = "Proceedings of the Tenth Conference on Machine Translation",
    month = nov,
    year = "2025",
    address = "Suzhou, China",
    publisher = "Association for Computational Linguistics",
    url = "https://aclanthology.org/2025.wmt-1.24/",
    doi = "10.18653/v1/2025.wmt-1.24",
    pages = "436--483",
    ISBN = "979-8-89176-341-8"
}

@inproceedings{blagec-etal-2022-global,
    title = "A global analysis of metrics used for measuring performance in natural language processing",
    author = "Blagec, Kathrin  and
      Dorffner, Georg  and
      Moradi, Milad  and
      Ott, Simon  and
      Samwald, Matthias",
    editor = "Shavrina, Tatiana  and
      Mikhailov, Vladislav  and
      Malykh, Valentin  and
      Artemova, Ekaterina  and
      Serikov, Oleg  and
      Protasov, Vitaly",
    booktitle = "Proceedings of NLP Power! The First Workshop on Efficient Benchmarking in NLP",
    month = may,
    year = "2022",
    address = "Dublin, Ireland",
    publisher = "Association for Computational Linguistics",
    url = "https://aclanthology.org/2022.nlppower-1.6/",
    doi = "10.18653/v1/2022.nlppower-1.6",
    pages = "52--63"
}

@Inbook{Way2018,
author="Way, Andy",
editor="Moorkens, Joss
and Castilho, Sheila
and Gaspari, Federico
and Doherty, Stephen",
title="Quality Expectations of Machine Translation",
bookTitle="Translation Quality Assessment: From Principles to Practice",
year="2018",
publisher="Springer International Publishing",
address="Cham",
pages="159--178",
isbn="978-3-319-91241-7",
doi="10.1007/978-3-319-91241-7_8",
url="https://doi.org/10.1007/978-3-319-91241-7_8"
}

@article{Terribile2024,
	Author = {Silvia Terribile},
	Journal = {Translation Spaces},
	Pages = {171--199},
	Title = { Is post-editing really faster than human translation? },
	Volume = {13},
    Number = {2},
	Year = {2024},
    doi = {https://doi.org/10.1075/ts.22044.ter}
}

@inbook{Rojo2017,
author = {Rojo, Ana},
publisher = {John Wiley \& Sons, Ltd},
isbn = {9781119241485},
title = {The Role of Creativity},
booktitle = {The Handbook of Translation and Cognition},
chapter = {19},
pages = {350-368},
doi = {https://doi.org/10.1002/9781119241485.ch19},
url = {https://onlinelibrary.wiley.com/doi/abs/10.1002/9781119241485.ch19},
eprint = {https://onlinelibrary.wiley.com/doi/pdf/10.1002/9781119241485.ch19},
year = {2017}
}

@article{Bassnettetal,
	Author = {Susan Bassnett and Lawrence Venuti and Jan Pedersen and Ivana Hostová},
	Journal = {World Literature Studies},
	Pages = {3--17},
	Title = {Translation and Creativity in the 21st Century.},
	Volume = {14},
    Number = {1},
	Year = {2022},
    doi = {https://doi.org/10.31577/WLS.2022.14.1.1}
}

@misc{lommel2024multirangetheorytranslationquality,
      title={The Multi-Range Theory of Translation Quality Measurement: MQM scoring models and Statistical Quality Control}, 
      author={Arle Lommel and Serge Gladkoff and Alan Melby and Sue Ellen Wright and Ingemar Strandvik and Katerina Gasova and Angelika Vaasa and Andy Benzo and Romina Marazzato Sparano and Monica Foresi and Johani Innis and Lifeng Han and Goran Nenadic},
      year={2024},
      eprint={2405.16969},
      archivePrefix={arXiv},
      primaryClass={cs.CL},
      url={https://arxiv.org/abs/2405.16969}, 
}

@misc{co2em,
    author = "Wong, Vinnie",
    year = "2024",
    title = {{Gen AI’s Environmental Ledger: A Closer Look at the Carbon Footprint of ChatGPT}},
    howpublished = {\url{https://piktochart.com/blog/carbon-footprint-of-chatgpt/}},
    note = {{Accessed: 2026/23/03}}
}

@misc{co2em-claude,
    author = "Christine Hall",
    year = "2025",
    title = {{What’s Your Chatbot’s Carbon Footprint?}},
    howpublished = {\url{https://fossforce.com/2025/04/whats-your-chatbots-carbon-footprint/}},
    note = {{Accessed: 2026/23/03}}
}

@inproceedings{FanFic_1,
    author = {Mia Jacobsen and Yuri Bizzoni and Pascale Feldkamp Moreira and Kristoffer L. Nielbo},
    title = {Patterns of Quality: Comparing Reader Reception Across Fanfiction and Commercially Published Literature} ,
    booktitle = {Proceedings of the Computational Humanities Research Conference 2024},
    pages = {718--739},        
    year = {2024},
}

@article{Fanfic_2,
    author = {Zhivar Sourati Hassan Zadeh and Nazanin Sabri and Houmaan Chamani and Behnam Bahrak },
	Journal = {Social Network Analysis and Mining},
	title = {Quantitative analysis of fanfictions’ popularity} ,
	Volume = {12},
    Number = {42},
	Year = {2021},
    doi = {https://doi.org/10.31577/WLS.2022.14.1.1}
}

@article{Alfassi04032026,
author = {Roi Alfassi and Angelora Cooper and Zoe Mitchell and Mary Calabro and Orit Shaer and Osnat Mokryn},
title = {Fanfiction in the Age of AI: Community Perspectives on Creativity, Authenticity and Adoption},
journal = {International Journal of Human–Computer Interaction},
volume = {42},
number = {5},
pages = {3062--3094},
year = {2026},
publisher = {Taylor \& Francis},
doi = {10.1080/10447318.2025.2531272},
URL = { 
    
        https://doi.org/10.1080/10447318.2025.2531272
    
},
eprint = { 
    
        https://doi.org/10.1080/10447318.2025.2531272
       

}
}

@inproceedings{du-etal-2025-optimising,
    title = "Optimising {C}hat{GPT} for creativity in literary translation: A case study from {E}nglish into {D}utch, {C}hinese, {C}atalan and {S}panish",
    author = "Du, Shuxiang  and
      Arenas, Ana Guerberof  and
      Toral, Antonio  and
      Gerrits, Kyo  and
      Borillo, Josep Marco",
    editor = "Bouillon, Pierrette  and
      Gerlach, Johanna  and
      Girletti, Sabrina  and
      Volkart, Lise  and
      Rubino, Raphael  and
      Sennrich, Rico  and
      Farinha, Ana C.  and
      Gaido, Marco  and
      Daems, Joke  and
      Kenny, Dorothy  and
      Moniz, Helena  and
      Szoc, Sara",
    booktitle = "Proceedings of Machine Translation Summit XX: Volume 1",
    month = jun,
    year = "2025",
    address = "Geneva, Switzerland",
    publisher = "European Association for Machine Translation",
    url = "https://aclanthology.org/2025.mtsummit-1.44/",
    pages = "578--591",
    ISBN = "978-2-9701897-0-1"
}

@misc{stureborg2024largelanguagemodelsinconsistent,
      title={Large Language Models are Inconsistent and Biased Evaluators}, 
      author={Rickard Stureborg and Dimitris Alikaniotis and Yoshi Suhara},
      year={2024},
      eprint={2405.01724},
      archivePrefix={arXiv},
      primaryClass={cs.CL},
      url={https://arxiv.org/abs/2405.01724}, 
}

@misc{Gaoetal,
      title={Detecting and Evaluating Bias in Large Language Models: Concepts, Methods, and Challenges}, 
      author={Zu Gao and Lingbo Tong and Zhiyong Zhana},
      year={2025},
      journal ={Journal of Behavioral Data Science},
     volume = {5},
    number ={2},
      url={https://jbds.isdsa.org/public/journals/1/html/v6n1/gao/},
    doi ={10.35566/jbds/gao}
}
\bibliographystyle{eamt26}

\appendix 

\section{Limitations}
\label{sec:limitations}
%Throughout the paper, we have mentioned limitations of our study. For clarity's sake, we also wanted to collect them in one section.
These are the limitations of this exploratory study.

\begin{itemize}
\item Our dataset is %relatively
small, as it consists of six source texts from 2 languages translated into three modalities for two languages (HT, PE, MT). This means we have 27 translations (3 modalities * 3 genres * 3 language pairs) in total.  
\item The texts %themselves
are also %relatively
small, as they are about 150 words each. Some of the AEMs and perhaps even LLM-as-a-judge might perform better overall on longer texts, especially if they do not score or judge on sentence-level but on document-level %the text as a whole
(such as LiTransProQA).
\item We only had a few %a handful of
human annotators available, which means the texts have been annotated by two to four people depending on the language pair. This makes it more difficult to see which annotations are more idiosyncratic than others and to see how the automatic metrics deal with those differences.
\item We evaluated LLM-as-a-judge on segment-level (per sentence), as document-level created very little output (see Appendix \ref{sec:Prompt} for details). However previous research has indicated this paradigm struggles more on segment-level compared to system-level (which 
we were limited to because of the dataset size) %we cannot do as we have limited texts, see above) \cite{moghe-etal-2023-extrinsic,freitag-etal-2023-results}.
\item We only used one LLM and one method for analysis. The model was SOTA (ChatGPT's GPT-5.2) and we did try out multiple other LLMs and methods to make sure our LLM-as-a-judge was of high quality, but it would be interesting to see how other models and prompting strategies compare to each other. % perform.

\end{itemize}

\section{Prompt template for the creation of the machine translation}
\label{sec:MT-template}
As mentioned in Section \ref{sec:Transl} we tried out different MT systems and prompts. Eventually we decided on a zero-shot English-language prompt to translate creatively, including information on author and the context of the fragment. We include the template of the prompts below using square brackets to indicate variable content such as authorial information and context. The individual prompts are included in our GitHub repository.\footnote{\url{https://github.com/INCREC/Creativity_bias}} \\
\newline 
\texttt{You are a professional translator. \\
Translate the following text from [source text] into [target text] creatively. It is a fictional text, a segment from a [genre] [title] by [author]. She/he is a(n) [country of origin] writer, [information on author and their style]. \\ This is a story about [content, including point of view, plot, and where the segment occurs in the narrative]. In the translation, please retain the style and the feel of the original. Please also explain your steps and reasoning throughout the process before giving me the final translation.}

\section{Counterbalancing of translations}
\label{sec:counterbalancingtransl}
The translators were counterbalanced across modalities, languages and genres for both translations and the annotations, shown in Table \ref{tab:counterb_transl}. 

\begin{table}[h]
    \centering
        \resizebox{0.85\columnwidth}{!}{
    \setlength{\tabcolsep}{4pt}
    \begin{tabular}{@{}cc|ccc@{}}
    \toprule
        \textbf{Language} & \textbf{Genre} & \textbf{Modality} & \textbf{Transl.}  & \textbf{Ann.}  \\ \midrule
     \multirow{3}{*}{EN-NL} & \multirow{3}{*}{Poem} & HT & T1 & T2  \\
     && PE & T2 & T1 \\
     && MT &  & T1\\      \midrule 
     \multirow{3}{*}{EN-NL} & \multirow{3}{*}{Short story} & HT & T2 & T1 \\
     & & PE & T1 & T2   \\
     & & MT & & T2 \\ \midrule
    \multirow{3}{*}{EN-NL} & \multirow{3}{*}{Thriller} & HT & T2 & T1 \\ 
     & & PE & T1 & T2  \\
     & & MT & & T2\\ \midrule
    \multirow{3}{*}{RU-NL} & \multirow{3}{*}{Poem} & HT & T2 & T1 \\
    && PE & T1 & T2 \\
    && MT && T2 \\ \midrule
    \multirow{3}{*}{RU-NL} & \multirow{3}{*}{Short story} & HT & T1 & T2 \\
    && PE & T2 & T1 \\
    && MT & & T1 \\ \midrule
     \multirow{3}{*}{RU-NL} & \multirow{3}{*}{Thriller} & HT & T1 & T2 \\
     && PE & T2 & T1 \\
     && MT & & T1 \\ \midrule
    \multirow{3}{*}{EN-CA} & \multirow{3}{*}{Poem} & HT & T3 & T4 \\
    && PE & T4 & T3 \\
    && MT && T3 \\ \midrule
    \multirow{3}{*}{EN-CA} & \multirow{3}{*}{Short story} & HT & T3 & T4 \\
    && PE & T4 & T3 \\
    && MT && T3 \\ \midrule
     \multirow{3}{*}{EN-CA} & \multirow{3}{*}{Thriller} & HT & T4 & T3 \\
     && PE & T3 & T4 \\
     && MT && T3 \\ 
 \bottomrule
    \end{tabular}
    }
\captionsetup{justification=centering, font=small}
    \caption{Overview of which translators translated (transl.) and annotated (ann.) which texts.}
    \label{tab:counterb_transl}
\end{table}

\begin{table}[h]
    \centering
    %\scriptsize
    \resizebox{\columnwidth}{!}{
    \setlength{\tabcolsep}{3pt}
    \begin{tabular}{@{}ccc|ccccc@{}}
    \toprule
    \textbf{Mod.} & \textbf{Lang} & \textbf{Genre} & \textbf{\# CS} & \textbf{\# Errors} & \textbf{\# EP} & \textbf{\# Kudos} & \textbf{CI} \\ \midrule
     HT & EN-NL & Poem & 12 & 3 & 3 & 3 & 50.0 \\
      PE & EN-NL & Poem & 8 & 8 & 8 & 4 & 31.0  \\
     MT & EN-NL & Poem & 5 & 16 & 48 & 1 & -6.3 \\
    \midrule
    HT & EN-NL & Short Story & 19 & 4& 8 & 3 & 80.0\\
     PE & EN-NL & Short Story & 10 & 5 & 5 &2 & 42.4 \\
    MT & EN-NL & Short Story & 2 & 17 & 109 & 1 & -48.2 \\
    \midrule
   HT & EN-NL & Thriller & 12 & 7 & 6 & 1 & 56.8 \\
     PE & EN-NL & Thriller & 5 & 11 & 18 & 0 & 13.3\\
    MT & EN-NL & Thriller & 5 & 9 & 29 & 1 & 6.8 \\
     \midrule
    HT & EN-CA & Poem & 12 & 9 & 9 &2 & 46.0 \\
    PE & EN-CA & Poem & 10 & 7 & 11 & 0 & 35.3 \\
    MT & EN-CA & Poem & 6 & 16 & 36 & 0 & 4.2 \\
     \midrule
    HT & EN-CA & Short Story & 7 & 3 & 3 & 3& 30.4 \\ 
    PE & EN-CA & Short Story & 3 & 7 & 19 & 2 & 4.1  \\
    MT & EN-CA & Short Story & 1 & 25 & 115 & 0 & -56.2 \\
     \midrule
    HT & EN-CA & Thriller & 3 & 4 & 4& 2 & 13.7\\
     PE & EN-CA & Thriller & 2 & 7 & 7 &1 & 6.1\\
    MT & EN-CA & Thriller & 0 & 18 & 66 & 0 & -42.9\\
        \midrule
    HT & RU-NL & Poem & 18 & 7 & 29 & 2& 44.7 \\
    PE & RU-NL & Poem & 16 & 5 & 23 & 0 & 40.8 \\
    MT & RU-NL & Poem & 7 & 14 & 176 & 0 & -149.8 \\
     \midrule
   HT & RU-NL & Short Story & 10 & 4 & 4 & 3 &  82.7\\
    PE & RU-NL & Short Story & 7 & 14 & 18 & 2 &48.4 \\
    MT & RU-NL & Short Story & 3 & 14 & 18 & 2 & 15.1 \\
     \midrule
    HT & RU-NL & Thriller & 10 & 9 & 9 & 0 & 52.9\\
    PE &RU-NL & Thriller & 7 & 14 & 18 &2 & 48.4\\
    MT & RU-NL & Thriller & 4 & 29 & 63 & 1 & -17.5 \\ 
 \bottomrule
    \end{tabular}
    }
    \captionsetup{justification=centering, font=small}
    \caption{Overview of human evaluation features (creative shifts (CS), errors, error points, Kudos and CI score) for each text. EP is short for Error Points. }
    \label{tab:humanscores} 
\end{table}

\section{Analysis of professional annotations}
\label{sec:detailed_human_ann}
\begin{figure}[h]
    \centering
    \includegraphics[scale=0.3]{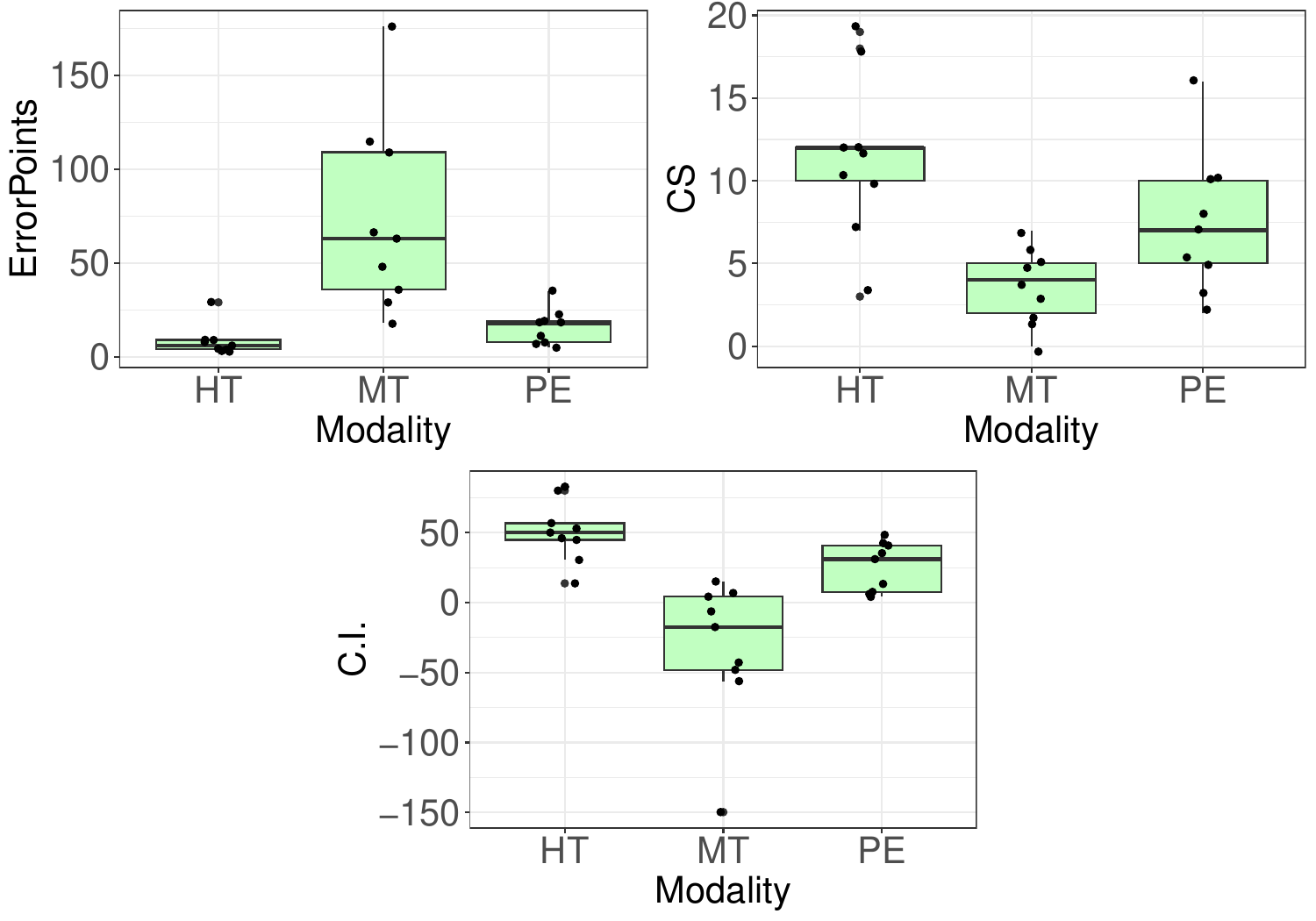}
    \caption{Box plots for errors, creative shift (CS) and creativity index (CI) per each modality level.}
    \label{fig:boxplot_mod}
\end{figure}

\begin{figure}[h]
    \centering
    \includegraphics[scale=0.3]{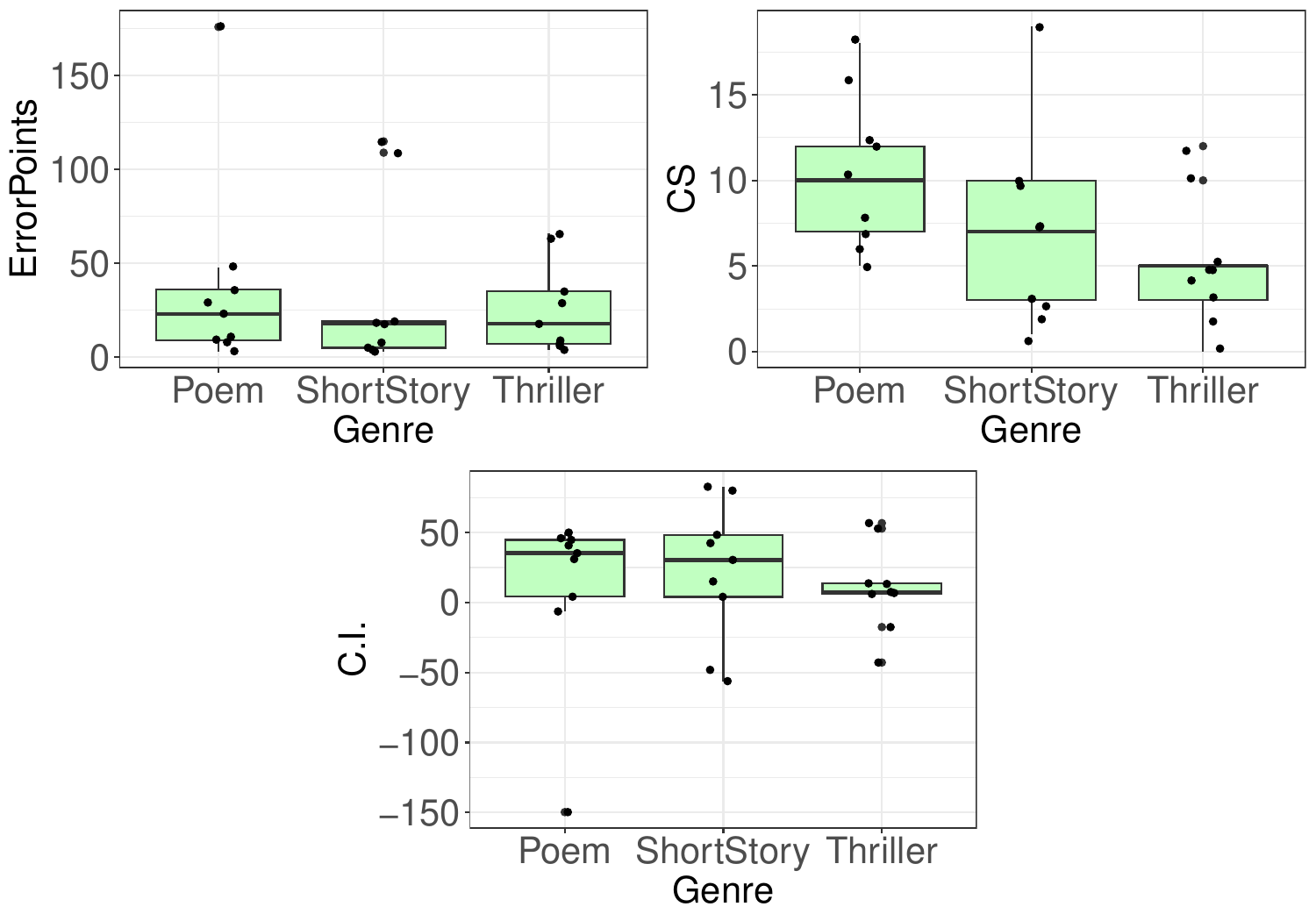}
    \caption{Box plots for errors, creative shift (CS) and creativity index (CI) per each genre level.}
    \label{fig:boxplot_genre}
\end{figure}

Section \ref{sec:hum_ann} briefly discussed the results from the professional annotations by discussing CI scores for the translated texts across modalities. %Here, we will show these professional annotations in more detail. 
Table \ref{tab:humanscores} shows the more detailed professional annotations for each text, including number of creative shifts, error points, Kudos and CI score. We created box plots for modality and genre across our human judgments variables, except for Kudos as this ranged only from 0 to 4, which did not result in informative box plots. These are shown in Figures \ref{fig:boxplot_mod} and \ref{fig:boxplot_genre}.

%As we were mainly interested in the effects of modality and genre w

%As discussed in the main body, we calculated the creativity index (CI) for all the translated texts. We used a Kruskal-Wallis rank sum test to see if any of the differences were significant (and we found that HT had a significantly higher creativity index (CI) than PE (p = .011) and MT ($p < .000$), with PE outperforming MT ($p < .000$). At the same time, MT had significantly more errors than HT (p = .002) and PE (p = .004), and PE significantly more than HT (p = .046). 

%We were mainly interested in potential differences across translation modalities and genres, so we created box plots for the different features of professional annotation (errors, CS and CI).  
%The plots show pronounced differences between modalities, with HT having fewer errors, more CSs and thus a higher CI, MT the complete opposite and PE in between both. Genre is less pronounced, but there too a difference between the high literary genre of the poem and the lower genre of the thriller with the literary short story in between. 
To test whether any of the differences were significant, we ran statistical tests. As there were relatively few texts we used non-parametric tests, specifically we used Kruskal-Wallis rank sum tests as the variables each have three levels. If this reaches significance, we run post-hoc comparisons using Wilcoxon Rank Sum tests with Holm-Bonferroni correction. The results of these tests are shown in Table \ref{tab:Statistical_test_human_ann}. %(There are significant differences between the modalities for all human judgment variables, only the differences in creative shifts between HT and PE and PE and MT is not significant. Fewer significant differences arise between the genres: only the differences on creative shifts is significant. Post-hoc tests show that there is a significant difference between the poem and thriller specifically. ) \todo{KG: I'm not completely sure about the final part (between brackets), can also be left out.} 

\begin{table}[h!]
 \scriptsize
    \centering
        \resizebox{0.8\columnwidth}{!}{
    \setlength{\tabcolsep}{4pt}
    \begin{tabular}{l |l c | lc | l c } 
    \toprule   
    & \multicolumn{2}{c|}{\textbf{Errors}} & \multicolumn{2}{c|}{\textbf{CS}} &\multicolumn{2}{c|}{\textbf{CI}} \\     & $\chi^2$ & p & $\chi^2$ & p & $\chi^2$ & p \\ \midrule
     \bf Modality & \bf 16.89 & \bf $<$ .000*** & \bf 11.28 & \bf .004** & \bf 17.60 & \bf $< $.000*** \\
    HT-MT & & $<$.000*** & & .008** & 
    & $<$.000*** \\ 
    HT-PE & & .011* & & .11 & & .011* \\
    PE-MT & & .004** && .11 & & .004**\\
    \midrule
    \bf Genre & \bf 0.53 & \bf.77& \bf 6.05 & \bf.049 & 0\bf .45 & \bf .79  \\
    Poem-Thriller && & & .04* && \\
    Thriller-ShortStory &&&& .26 && \\
    ShortStory-Thriller&&& & .69 & \\
    \bottomrule
    \end{tabular}
    }
     \captionsetup{justification = centering}
    \caption{Number of errors from professionals and LLM-as-a-judge (including matches) across genre.\\ ***p $<$.001, **p $<$.01, *p $<$.05}
    \label{tab:Statistical_test_human_ann}
\end{table}

\section{Overview of LLM-as-a-judge prompting strategies} 
\label{sec:Prompt}
%Here too not sure whether this is the right term or if we should use something else
As mentioned in the main body, we experimented with different prompting strategies across different models (see Section \ref{sec:AEMs}). We did not run systematic analyses of all texts across all strategies and models as this was not the main aim of the paper and it would have led to an (unnecessary) increase in costs and environmental impact, but we did try out multiple models and prompts using selected texts. Specifically, we tried out TSA \cite{yeom-etal-2025-tagged}, AutoMQM \cite{fernandes-etal-2023-devil}, GEMBA-MQM \cite{kocmi-federmann-2023-gemba} and EAPrompt \cite{lu-etal-2024-error}.

First, we tried out different strategies for the texts at document-level, as studies have shown these strategies tend to perform better at document-level than segment-level \cite{moghe-etal-2023-extrinsic,freitag-etal-2023-results}. However, this created outputs of only about 1 to 6 errors on MT-texts, whereas the professional annotation found on average 18 errors per MT-text. These errors also did not overlap, meaning it missed all professionally-marked errors and included non-errors. %For instance, in the MT of the EN-NL Short Story, AutoMQM only found one error, GEMBA-MQM five and TSA zero (whereas the human annotators had 17 errors). Furthermore, not one of those found errors overlapped with human annotators, and major or critical errors annotated by professional literary translators were not included in the model output. Even worse, one of the error of GEMBA-MQM and AutoMQM concerned a segment that the professional annotators had marked as Kudos. 

We thus decided to annotate on \textbf{segment-level (per sentence)}. This increased the number of errors included in each annotation. However, it also caused multiple instances of omission/addition when segments where only moved across sentences. To mitigate this we \textbf{included preceding and subsequent sentence as context} to the prompt as well as extra instructions in the prompt about omissions.%\footnote{Specifically, we specified not treating cross-sentence shifts as errors, if a phrase is moved to the previous/next sentence but translated correctly, to not label it as omission or addition, to only mark an omission/addition if the content is missing entirely or appears incorrectly in the translation as a whole, not just in this segment, and lastly to use the provided previous/next sentences to verify whether the content is simply moved rather than missing.} 
Although not all instances were solved, this did decrease the (superfluous) instances of omissions and deletions overall and was thus kept for the final prompt.
%However, now these annotations contained multiple instances of omission/addition when a sentence fragment was moved across sentences (e.g. “he asked” moved from before the question to after the question which shifted the sentence order). To mitigate these superfluous omissions and additions preceding and subsequent sentence were added for each sentence as context and—-when solely adding context did not change the output—-we included extra instructions about omissions in the prompt. Specifically, we specified not treating cross-sentence shifts as errors, if a phrase is moved to the previous/next sentence but translated correctly, to not label it as omission or addition, to only mark an omission/addition if the content is missing entirely or appears incorrectly in the translation as a whole, not just in this segment, and lastly to use the provided previous/next sentences to verify whether the content is simply moved rather than missing. Although not all instances were solved, this did decrease the (superfluous) instances of omissions and deletions overall and was thus kept for the final prompt.\AGA{This is a bit chaotic, and we do not want to give that impression. I think if here there was not a systematic approach, perhaps explain the few prompting strategies tested and then the ones you selected for a better outcome?}
After careful consideration, \textbf{Tagged Span Annotation (TSA)} by Yeom et al. \shortcite{yeom-etal-2025-tagged} was chosen as the prompt basis as this returned the most accurate results for our dataset among the existing prompts. 

For the model, we experimented with multiple models, including both reasoning models and non-reasoning ones. We focused on ones that are consistently used in literary translation according to the literature \cite{liu-etal-2023-g,yeom-etal-2025-tagged,kocmi-federmann-2023-gemba}. We found that \textbf{reasoning models} worked better than non-reasoning ones,\footnote{Non-reasoning models such as Claude's Sonnet 4.6 and  GPT-5.2 did not follow instructions consistently and also included error segments that were not even present in the text. %For instance, in the MT version of the EN-NL Thriller the text read \textit{naar de plek tussen de kapstok en de keuken} (to the space between the coat rack and the kitchen), but the error output marked \textit{naar de kapstok en keuken} (to the coat rack and the kitchen) with the explanation that it had indicated the wrong place and it should be \textit{naar de plek tussen de kapstok en de keuken} as it had been originally).
} and from the reasoning models \textbf{GPT-5.2} was the best performing on errors as Gemini 3 and Claude 3.7 Sonnet show erratic behaviour when marking errors. 

Lastly, we also tried out different levels of reasoning effort: increasing the effort generally improved performance, but this flattened out towards higher efforts: xhigh performed similarly to high but had much higher costs (both pecuniary and time-wise). We therefore decided to set the \textbf{reasoning effort to high}, which was used for all evaluations.

\section{Template prompt for LLM-as-a-judge evaluation}
\label{sec:LLM_template}
\subsection{Error evaluation prompt}
For the error evaluation we eventually settled on an adapted version of the TSA \cite{yeom-etal-2025-tagged} as described above. The entire code can be found on our GitHub repository. The prompt looked at follows: 
\newline

\texttt{You are a careful and balanced annotator for machine translation quality.
Your task is to identify translation errors with appropriate confidence. \\
\newline
\#\# EVALUATION GUIDELINES: \\
- Be thorough but precise \\
- Only mark errors when confident \\
- Focus on objective errors, not preferences \\
- Mark minimal error spans with $<$v0$>$, $<$v1$>,$ ... \\
- Do NOT treat cross-sentence shifts as errors. \\
- If a phrase is moved to the previous/next sentence but translated correctly, do NOT label it as omission or addition. \\
- Only mark an omission/addition if the content is missing entirely or appears incorrectly in the translation as a whole, not just in this segment. \\
- Use the provided previous/next sentences to verify whether the content is simply moved rather than missing. \\
\newline
\#\# ERROR CATEGORIES: \\
- Accuracy (addition, omission, mistranslation, overtranslation, undertranslation untranslated) \\
- Linguistic Convention (grammar, punctuation, spelling) \\
- Style (awkward, inconsistent, register, unidiomatic, audience appropriateness) \\
- Other \\
\newline
\#\# Severity: \\
- Major \\
- Minor \\
\newline
\#\# CRITICAL RULES: \\
- Tags must be sequential: $<$v0$>$, $<$v1$>$, $<$v2$>$... \\
- No explanations, comments, or extra text \\
- If omission: insert empty tags$ <$vN$>$$<$/vN$>$ \\
\newline
\#\#\# CONTEXT (NOT FOR EVALUATION): \\
Previous source: \{prev\_source\} \\
Previous translation: \{prev\_translation\}  \\
\newline
Next source: \{next\_source\} \\
Next translation: \{next\_translation\} \\
\newline
\#\#\# SOURCE TEXT (CURRENT SENTENCE): \\
\{source\} \\
\newline
\#\#\# TRANSLATION (CURRENT SENTENCE): \\
\{translation\}}

\subsection{Creative shift evaluation prompt}
Based on the TSA strategy for errors, we created a new prompt for creative shifts shown below. \\ 
\newline
\texttt{You are a careful and balanced annotator for creativity in translation.
Your task is to categorise potential creative segments (UCP) as a creative shift (CS), Reproduction (R), omission (O), or error (E).\\
\newline
\#\# EVALUATION GUIDELINES: \\
- Be thorough but precise\\
- Consider whether there is a direct or coined translation available \\}

\texttt{\#\# CREATIVITY CATEGORIES: \\
- Creative shift (CS) (All translations that deviate from the source with a different idea or image are considered creative shifts. These are the creative shifts (“non-literal”).) includes: \\
-- CSA (Abstraction): refers to those cases in which translators use more vague, general or abstract in the translation \\
-- CSM (Modification): refers to shifts that are at the same level of abstraction (e.g. express a source text metaphor with a different metaphor without the image becoming more abstract or concrete). In other words, the translation is modified for the target culture. \\
-- CSC (Concretisation): refers to instances when the translation evokes a more explicit, more detailed and more precise idea or image than the source text
- Reproduction (R): translations that reproduce the source text with the same idea or image, even if they are acceptable, are not considered a creative shift in the translation, but a reproduction. \\
- Omission (O): If a term or expression in the source text is omitted in the translation this will be marked as omission. An omission could correspond to a) creative solution (for example, that text was omitted because it does not make sense in the translation) or b) a shortcut solution (for example, that text is omitted because it is rather cumbersome to render). \\
- Error (E): If the translation is not acceptable (contains too many errors). \\
\newline
\#\# CRITICAL RULES: \\
- No explanations, comments, or extra text}

\section{Detailed scores for AEMs and LLM-as-a-judge}
\label{sec:More_details_scores}
\subsection{Detailed AEM scores}
\label{sec:AEMscoredetails}
The results of the AEM scores were briefly discussed in Section \ref{sec:results_AEMs} %before we continued to look at the correlations they had with professional scores.
More detailed analysis of AEMs will be discussed here. The scores are shown in Table \ref{tab:AEMscores} below.

\subsection{Professional vs LLM-as-a-judge for creativity}
\label{sec:app_proportions_crea}

Table \ref{tab:CreativityDistribution} shows the distribution of creative shifts (CS) and reproductions (R) across the modalities for the professional and the LLM-as-a-judge annotations. %What is immediately apparent is that LLM-as-a-judge assigns almost the same number of CSs for each modality, whereas there is a clear difference between the modalities in the professional annotations (with HT assigned most CSs, followed by PE and then MT). This means that to HT and PE, LLM-as-a-judge assigns fewer CSs than professional annotations do, but to MT LLM-as-a-judge assigns more. 
Looking at the proportions of matching assignments by professionals and LLM-as-a-judge, we see that LLM-as-a-judge performs best in MT (128 / 188, 68\%), closely followed by PE (121 / 188, 64\%) with HT trailing slightly further behind (110 / 188, 59\%). This seems largely due to the reproductions (110 of the 127 matches in MT are for reproductions for instance), but perhaps this is another indication that automatic metrics such as LLM-as-a-judge performs better on MT than on other modalities.

\subsection{CI-scores for professionals and LLM-as-a-judge}
\label{sec:CI-scores}
Table \ref{tab:CI_scores_comb} shows the CI scores for each text from professionals and LLM-as-a-judge per language, genre and modality, showing the lack of correlations between professionals and LLM-as-a-judge. %discussed in Section \ref{sec:LLM-as-a-judge}. %The table shows the lack of correlations between professionals and LLM-as-a-judge discussed in Section \ref{sec:LLM-as-a-judge}. For professional annotators, the CI is indicative of translation quality with big gaps between the modalities throughout, showing a pronounced difference between the translation modalities. For LLM-as-a-judge this gap is minimalised or even completely removed from the CI scores. Related to this, we also see a much smaller range of CI scores from the model (between 31.75 and -52.93 vs 82.71 and -149.78 for professionals). This further highlights the flattening that occurs for CI scores when calculated by LLM-as-a-judge. This means LLM-generated CI scores do not accurately reflect differences in creativity and efface creative solutions, which suggests LLM-generated CI are unreliable as indicators of quality.

\begin{table}[h]
    \centering
        \resizebox{0.8\columnwidth}{!}{
    \begin{tabular}{@{}ccc|ccc|c@{}}
    \toprule
       &&&  \multicolumn{3}{c}{\textbf{Professional}} \\ \midrule
        && & CS & R & O &  Total \\ \midrule
     \multirow{14}{*}{\textbf{LLM}}&  \multirow{3}{*}{HT} & CS & 41 & 12 & 2&55 \\
     & & R & 59 & 68 &0 & 127 \\
     && O & 3 & 2 & 1 & 6\\ \cmidrule{2-7}
      && Total & 103 & 82 & 3& 188\\ \midrule
    & \multirow{3}{*}{PE} & CS & 32 & 24 & 0 & 56 \\
   && R & 34 & 87 & 1 & 122 \\
   && O & 0 & 8 & 2 & 10\\ \cmidrule{2-7}
 && Total & 66 & 119& 3 & 188\\ \midrule
& \multirow{3}{*}{MT} & CS & 17 & 37 & 0&54 \\
&& R & 17 &110 &1 & 128 \\
&& O & 1 & 4 & 1 & 6\\ \cmidrule{2-7}
&& Total & 35 & 151 & 2 & 188 \\ \bottomrule
    \end{tabular}
    }
    \captionsetup{justification = centering, font=small}
    \caption{Distribution of creativity classification for professionals and LLM-as-a-judge per modality.}
    \label{tab:CreativityDistribution}
\end{table}

\begin{table}[h]
    \centering
    \resizebox{0.85\columnwidth}{!}{
    \begin{tabular}{ccc|ccc} 
    \toprule
\bf Lang & \bf Genre & \bf \bf Mod. & \bf Human &\bf LLM \\ \midrule
    \multirow{9}{*}{EN-NL} & \multirow{3}{*}{Poem} & HT & 50.0 & 24.7 \\ 
    & & PE & 31.0& 22.2 \\
    & & MT & -6.3 & 17.6\\ \cmidrule{2-5}
    & \multirow{3}{*}{Short Story} & HT & 80.0 & 16.8 \\
    & &PE & 42.4 & 19.4\\
    & &MT & -48.2 & -6.2\\ \cmidrule{2-5}
    & \multirow{3}{*}{Thriller} & HT & 56.8 & 27.0 \\
    && PE & 13.3 & 24.0 \\
    && MT & 6.8 & 13.6 \\
        \midrule
        \multirow{9}{*}{EN-CA} & \multirow{3}{*}{Poem} & HT & 46.0 & 26.8 \\
        && PE & 35.3 & 25.0 \\
        && MT & 4.2& 12.4 \\ \cmidrule{2-5}
        & \multirow{3}{*}{Short Story} & HT & 30.4 & 3.8 \\
        && PE & 4.1 & 7.8 \\
        && MT & -56.2 & -2.0 \\ \cmidrule{2-5}
        & \multirow{3}{*}{Thriller} & HT & 13.7 & 18.3 \\
        & & PE & 6.1 & 2.0 \\
        && MT & -42.9 & 27.2 \\ \midrule
        \multirow{9}{*}{RU-NL} & \multirow{3}{*}{Poem} & HT & 44.7 & -11.4 \\
        && PE & 40.8 & -52.9 \\
        && MT & -149.8 & -26.8 \\ \cmidrule{2-5}
        & \multirow{3}{*}{Short Story}& HT & 82.7 & 3.0 \\
        && PE & 48.4 &15.1 \\
        && MT & 15.1 & 11.3 \\ \cmidrule{2-5}
        & \multirow{3}{*}{Thriller} & HT & 52.9 & 12.3 \\
        && PE & 7.6 & 24.6 \\
        && MT & -17.5 & 31.8\\ 
    \bottomrule
    \end{tabular}
   }
     \captionsetup{justification = centering}
    \caption{CI scores from professional and LLM-as-a-judge evaluation for each text.}
    \label{tab:CI_scores_comb}
\end{table}

\begin{table*}[h]
     \resizebox{\textwidth}{!}{
    \setlength{\tabcolsep}{3pt}
    \centering
    \begin{tabular}{@{}ccc|cccccccccc@{}}
    \toprule
    \textbf{Mod.} &  \textbf{Lang} &  \textbf{Genre} &  \textbf{BERTscore$\uparrow$} &  \textbf{BLEU$\uparrow$} &  \textbf{BLEURT$\uparrow$} &  \textbf{chrF2$\uparrow$} &  \textbf{COMET$\uparrow$} &  \textbf{COMETKiwi$\uparrow$} &  \textbf{LiTransProQA$\uparrow$} &  \textbf{MetricX24$\downarrow$} &  \textbf{TER$\downarrow$}           \\ \midrule
     HT & EN-NL & Poem & & & && & 78.0 & \bf 22.6 & & \\
      PE & EN-NL & Poem & \bf 89.3 & \bf38.1 & \bf 71.4 & \bf 59.6 & \bf 83.0 & \bf 78.6 & 21.7 & \bf 2.80 & \bf 39.8 \\
     MT & EN-NL & Poem & 86.4 & 29.9 & 69.3 & 55.7 & 80.0 & 75.3 & \bf 22.6 & 2.81 & 45.4 \\
    \midrule
    HT & EN-NL & Short Story & & && & & 76.6 & \bf 22.6 && \\
     PE & EN-NL & Short Story & \bf 85.5 & \bf 37.6 & \bf 69.8 & \bf 55.5 & \bf 83.5 & 82.1 & 19.3 & \bf 2.22 & \bf 62.3 \\
    MT & EN-NL & Short Story & 81.2 & 27.7 & 66.4 & 50.2 & 82.0 & \bf 82.2 & 21.7 & 2.46 & \bf 62.3\\
    \midrule
   HT & EN-NL & Thriller & & && & & 75.5 & \bf 21.6 & & \\
     PE & EN-NL & Thriller & \bf 86.8 & \bf 37.6 & \bf 70.3 & \bf 57.9 & \bf 80.8 & \bf 79.4 & 16.7 & \bf 2.95 & \bf 49.1 \\
    MT & EN-NL & Thriller & 85.0 & 36.5 & 68.3 & 56.6 & 77.9 & 75.2 & 19.9 & 3.47 & 49.7 \\ 
     \midrule
    HT & EN-CA & Poem & & & && & 58.5 & \bf 22.6 && \\
    PE & EN-CA & Poem & 79.8 & 17.8 & 44.3 & \bf 40.3 & 68.0 & 60.8 & 22.1 & 7.14 & \bf 76.4\\
    MT & EN-CA & Poem & \bf 81.1 & \bf 18.6 & \bf 49.0 & 39.9 & \bf 70.6 & \bf 69.7 & 22.1 & \bf 6.21 & 77.7 \\
     \midrule
    HT & EN-CA & Short Story & & & & & & 70.9 & 20.3 & & \\ 
    PE & EN-CA & Short Story & \bf 85.1 & \bf 29.8 & \bf 69.3 & \bf 52.3 & \bf 81.8 & 72.1 & 22.1 & \bf 3.72 & \bf 56.5 \\
    MT & EN-CA & Short Story & 83.8 & 24.7 & 61.5 & 47.3 & 81.0 & \bf  77.4 & \bf 22.6 & 4.33 & 63.5 \\
     \midrule
    HT & EN-CA & Thriller && & & & & 61.3 & 1.0 && \\
     PE & EN-CA & Thriller & 73.1 & 0.4 & 30.3 & \bf 18.1 & 52.9 & 63.7 & \bf 22.6 & 5.57 & 111.0 \\
    MT & EN-CA & Thriller & \bf 74.8 & \bf 0.5 & \bf 34.3 & 17.8 & \bf 54.8 & \bf 67.1 & 18.9 & \bf 4.53 & \bf 109.7\\
        \midrule
    HT & RU-NL & Poem & && & & & 50.2 & 16.8 && \\
    PE & RU-NL & Poem & 58.3 & \bf 7.2 & 24.6 & 20.0 & 59.9 & 47.4 & 15.8 & 9.22 & \bf 101.8 \\
    MT & RU-NL & Poem & \bf 59.5 & 2.7 & \bf 28.1 & \bf 23.0 & \bf 63.8 & \bf 54.5 & \bf 17.6 & \bf 6.78 & 110.9 \\
     \midrule
   HT & RU-NL & Short Story & & & && & \bf 73.8 & 21.6 && \\
    PE & RU-NL & Short Story & 76.9 & 21.8 & 68.6 & \bf 53.5 & 77.9 & 71.8 & \bf22.6 & 3.14 &62.1 \\
    MT & RU-NL & Short Story & \bf 78.3 & \bf 25.2 & \bf 72.4 & \bf 53.5 & \bf 78.9 & 70.3 & 21.7 & \bf 2.43 & \bf 59.1 \\
     \midrule
    HT & RU-NL & Thriller & & && & & \bf 71.2 & 21.7 && \\
    PE &RU-NL & Thriller & \bf 69.7 & \bf 31.9 & \bf 55.3 & \bf 53.8 & \bf 80.3 & 67.0 & 21.7 & \bf 2.62 & \bf 51.4 \\
    MT & RU-NL & Thriller & 66.5 & 15.5 & 50.3 & 43.4 & 77.1 & 66.7 & \bf 22.6 & 3.32 & 68.5 \\
 \bottomrule
    \end{tabular}
    }
    \captionsetup{justification=centering, font=small}
    \caption{Overview of AEM scores for all translations in the text.}
    \label{tab:AEMscores} 
\end{table*}

\end{document}